\documentclass{article}

\usepackage[eandd, final]{neurips_2026}
\usepackage{microtype}
\usepackage{graphicx}
\usepackage{subcaption}
\usepackage{booktabs} % for professional tables
\usepackage{colortbl}
\usepackage{graphicx}
\usepackage{booktabs}
\usepackage{amsmath} 
\usepackage{amssymb}
\usepackage{mathtools}
\usepackage{amsthm}
\usepackage{multirow,makecell}
\usepackage{multicol}
\usepackage{mathrsfs}
\usepackage{diagbox}
\usepackage{dsfont}
\usepackage{hyperref}
\usepackage{xurl} 
\usepackage{enumitem}
\usepackage{pifont}% http://ctan.org/pkg/pifont
\usepackage{tcolorbox}
\tcbuselibrary{breakable, skins} % <--- 加上这两个库
\usepackage{xcolor}
\usepackage{listings}

\usepackage[utf8]{inputenc} % allow utf-8 input
\usepackage[T1]{fontenc}    % use 8-bit T1 fonts
\usepackage{hyperref}       % hyperlinks
\usepackage{url}            % simple URL typesetting
\usepackage{booktabs}       % professional-quality tables
\usepackage{amsfonts}       % blackboard math symbols
\usepackage{nicefrac}       % compact symbols for 1/2, etc.
\usepackage{microtype}      % microtypography
\usepackage{xcolor}         % colors
\usepackage{wrapfig}

\def\bench{\textrm{CineMME}} 
\def\tracka{\textrm{CineMME-Grounding}} 
\def\trackb{\textrm{CineMME-Reasoning}} 
\newcommand{\tabtitlecolor}{\rowcolor{blue!6}}
\newcommand{\cmark}{\textcolor{darkgreen}{\ding{51}}}
\newcommand{\xmark}{\textcolor{red}{\ding{55}}}

\newcommand{\violet}[1]{\textcolor{violet}{#1}}

\def\ie{\textit{i.e.}}

\definecolor{darkgreen}{RGB}{0,150,0}  % Dark green
\definecolor{darkred}{RGB}{255,0,0} 
\usepackage{hyperref}

\usepackage{amsthm}

\definecolor{MyDarkRed}{rgb}{0.8,0.02,0.02}
\definecolor{MyDarkBlue}{rgb}{0.02,0.02,0.8}
\definecolor{MyDarkGreen}{rgb}{0.1,0.8,0.1}
\definecolor{darkgreen}{rgb}{0.0, 0.5, 0.0}

\usepackage[capitalize,noabbrev]{cleveref}

\theoremstyle{plain}

\theoremstyle{definition}

\theoremstyle{remark}

\usepackage[disable,textsize=tiny]{todonotes}
\title{

CineMME: From Dialogue Grounding to Narrative Reasoning in Multimodal Video Understanding
}

\author{
\vspace{0.12cm}
Jin~Liu$^{1,2,4}$\hspace{0.15cm}
Dawei~Du$^{4*}$\hspace{0.15cm}
Yexiang~Liu$^{2,3,4}$\hspace{0.15cm}
Sijie~Zhu$^{4}$\hspace{0.15cm}
Fan~Chen$^{4}$\\[3pt]
\bfseries Junxian~Duan$^{2,3}$\hspace{0.15cm}
Huaibo~Huang$^{2,3}$\hspace{0.15cm}
Ran~He$^{1,2,3}$\hspace{0.15cm}
Longyin~Wen$^{4}$\hspace{0.15cm}
Zhenfang~Chen$^{4}$\thanks{Corresponding authors. Emails: {dawei.du@bytedance.com,zhenfang.chen@bytedance.com}}\\[4pt]
$^1$School of Information Science and Technology, ShanghaiTech University, Shanghai, China \\
$^2$MAIS \& NLPR, Institute of Automation, Chinese Academy of Sciences, Beijing, China \\
$^3$School of Artificial Intelligence, University of Chinese Academy of Sciences, Beijing, China  \\
$^4$ ByteDance, San Jose, USA.
}

\begin{document}

\maketitle

\begin{center}
        \centering
        \vspace{-20pt} 
        {\small Project page: \url{https://cinemme.github.io/}\par}
        \vspace{6pt}
        \includegraphics[width=1\textwidth]{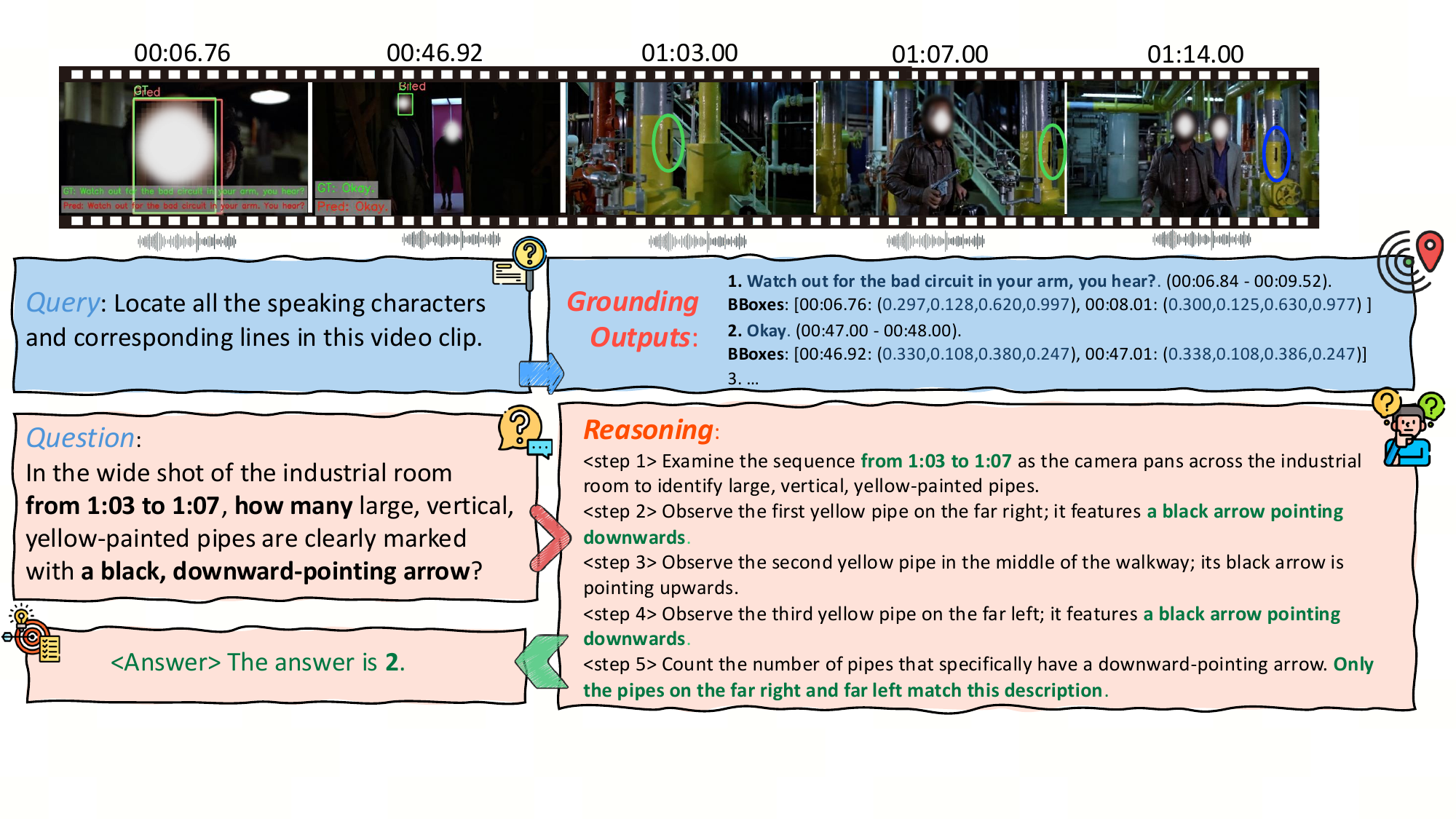}
        \vspace{-30pt}
        
        \captionof{figure}{
            Examples from CineMME.
            \textbf{Top:} 
            \tracka~evaluates dialogue-level evidence acquisition, requiring models to localize when speech occurs, transcribe what is said, and ground the active speaker to the corresponding speaker face.
            \textbf{Bottom:} 
            \trackb~evaluates whether models can use temporally grounded evidence to answer open-ended narrative questions.
            The two tracks are linked by design, enabling diagnosis of how speech grounding and speaker binding support plot-centric reasoning.
        }
        \label{fig:teaser}
    \end{center}

\begin{abstract}
Narrative video understanding requires more than recognizing visible events: models must extract evidence from dialogue, bind it to speakers, and use it to infer narrative meaning.
Existing video benchmarks evaluate aspects of this problem, including video question answering, audio-visual reasoning, and temporal grounding, but rarely test whether these abilities form a coherent dialogue-to-plot evidence chain within the same clips.
We introduce \textbf{CineMME}, a human-verified benchmark designed to diagnose dialogue-grounded plot understanding, supported by approximately $1,600$ expert person-hours of annotation.
CineMME contains $975$ curated short narrative video clips and two linked tracks: \textit{CineMME-Grounding}, with $13,554$ timestamped speech segments and $33,083$ speaker face boxes for temporal dialogue localization, transcription, and active-speaker grounding; and \textit{CineMME-Reasoning}, with $1,600$ multiple-choice and $316$ open-ended questions spanning perception, audio, narrative, social, and scene-presentation understanding.
Our main evaluation compares $20$ MLLMs on reasoning and $9$ omni-modal models on grounding.
The results reveal a pronounced who-what-when decoupling: Gemini-3 Pro localizes dialogue in time ($68.34\%$ tIoU) and transcribes it with a WER of $28.99\%$, yet achieves only $7.74\%$ vIoU for active-speaker grounding while achieving $67.66\%$ MCQ accuracy on questions with extractable answers.
Modality ablations further show that vision is not uniformly beneficial: it improves performance on perception-heavy questions but can hurt dialogue-centric reasoning.
CineMME provides a compact diagnostic testbed for evaluating how multimodal video models connect dialogue, speakers, visual evidence, and narrative meaning.
\end{abstract}

\section{Introduction}

Narrative video understanding is not merely a problem of recognizing visible events.
Narrative clips convey meaning through the interaction of dialogue, speaker identity, facial expressions, off-screen voices, sound, editing, visual composition, and plot context.
A model may need to determine \textit{when} a line is spoken, \textit{what} is said, \textit{who} says it, and why the utterance matters for character relationships or story development.
In this setting, reasoning is closely tied to evidence acquisition: a correct narrative inference may depend on a short utterance, an off-screen speaker, a reaction shot, or a mismatch between what is said and what is visually shown.
This makes narrative videos a natural stress test for multimodal large language models (MLLMs) that aim to integrate vision, audio, speech, and language over time.

Existing video benchmarks have made substantial progress in long-video understanding, audio-visual reasoning, movie question answering (QA), and spatio-temporal grounding~\cite{fuVideoMMEFirstEverComprehensive2025,rawalCinePileLongVideo2024,tapaswiMovieQAUnderstandingStories2016,yueMovie101NewMovie2023,gu2024context,yueMovie101v2ImprovedMovie2025,gengLongVALEVisionAudioLanguageEventBenchmark2025,gongAVOdysseyBenchCan2024,zhangTimeLensRethinkingVideo2025,hannanSVAGBenchLargeScaleBenchmark2025,gaoOmniGroundComprehensiveSpatioTemporal2025}.
However, they typically measure either final-answer correctness or isolated localization ability, rather than whether a model acquires and binds the evidence needed for plot-centric reasoning.
As a result, a model may answer a plot question correctly while failing to localize the relevant utterance, identify the active speaker, or integrate speech with visual and narrative context.
What is missing is therefore a diagnostic evaluation of the dialogue-to-plot evidence chain: \emph{when} dialogue occurs, \emph{what} is said, \emph{who} says it, and \emph{why} it matters.

To fill this gap, we introduce \textbf{CineMME}, a human-verified benchmark for diagnosing dialogue-grounded plot understanding.
CineMME contains $975$ short narrative video clips, ranging from $22.1$ to $300.1$ seconds in duration and spanning diverse genres.
CineMME is organized around two linked tracks that correspond to different stages of the same evidence chain.
\textit{CineMME-Grounding} evaluates evidence acquisition and binding, with $546$ videos, $13,554$ timestamped speech segments, and $33,083$ speaker face boxes for temporal dialogue localization, transcription, and active-speaker grounding.
\textit{CineMME-Reasoning} evaluates evidence use, with $667$ videos, $1,600$ multiple-choice questions, and $316$ open-ended questions spanning \textit{Perception}, \textit{Audio}, \textit{Narrative}, \textit{Social}, and \textit{Scene Presentation} understanding.
The $238$ clips shared by the two tracks enable clip-level analysis of whether models that answer plot questions correctly can also acquire and bind the multimodal evidence relevant to those answers.

CineMME is constructed through a semi-automatic, human-verified pipeline designed to support both scalability and diagnostic reliability.
We first use WhisperX~\cite{bainWhisperXTimeAccurateSpeech2023} and InsightFace~\cite{deng2018arcface} to generate initial speech segments and face-track metadata.
MLLMs are then used to propose speaker attributions, candidate questions, and answer choices, followed by adversarial distractor refinement to reduce shortcut-based answers.
All final labels are determined through human annotation followed by independent review.
Agreement on paired, non-missing verification decisions between annotation and review stages covers ASR, timestamps, speaker IDs, and discarding, with Cohen's $\kappa$ ranging from $0.89$ to $0.94$.
These values summarize consistency across sequential review stages.
This design makes CineMME suitable not only for ranking models, but also for analyzing where model failures arise along the evidence chain.

We evaluate $20$ MLLMs~\cite{microsoftPhi4MiniTechnicalReport2025,teamKimiVLTechnicalReport2025,wangInternVL35AdvancingOpenSource2025,liuMinistral32026,liXiaomiMiMoVLMilocoTechnical2025,teamGLM45VGLM41VThinkingVersatile2025,baiQwen3VLTechnicalReport2025,baiQwen25VLTechnicalReport2025,yaoMiniCPMVGPT4VLevel2024,yeOmniVinciEnhancingArchitecture2025,xuQwen25OmniTechnicalReport2025,xuQwen3OmniTechnicalReport2025,cui2026minicpmo45realtimefullduplex,OpenAI2025GPT5SystemCard,xiaomi2026mimo25,qwenteam2026qwen35omnitechnicalreport,gemini-3-pro} on reasoning and $9$ omni-modal models~\cite{yaoMiniCPMVGPT4VLevel2024,yeOmniVinciEnhancingArchitecture2025,microsoftPhi4MiniTechnicalReport2025,xuQwen25OmniTechnicalReport2025,xuQwen3OmniTechnicalReport2025,cui2026minicpmo45realtimefullduplex,xiaomi2026mimo25,gemini-3-pro,qwenteam2026qwen35omnitechnicalreport} on grounding.
Our experiments are designed to answer three diagnostic questions:
1) Can current models perform high-level plot-centric reasoning?
2) Can they acquire and bind dialogue-level evidence through temporal localization, transcription, and active-speaker grounding?
3) Does better grounding correspond to stronger downstream reasoning on shared clips?
The results reveal weaknesses at multiple stages of the dialogue-to-plot evidence chain, beyond final question answering.
On questions with extractable answers, Gemini-3 Pro achieves the highest MCQ accuracy among the evaluated models at $67.66\%$, while MiMo-v2.5 achieves the highest accuracy among open-weight models at $58.44\%$.
The results reveal a pronounced who-what-when decoupling: Gemini-3 Pro localizes dialogue in time ($68.34\%$ tIoU) and transcribes it with a WER of $28.99\%$, yet achieves only $7.74\%$ vIoU for active-speaker grounding.
On shared clips, temporal localization and transcription quality are positively associated with open-ended question (OEQ) scores in the reported splits, while associations between spatial grounding and OEQ scores are mixed.
Finally, modality ablations reveal that visual input is selectively useful rather than universally beneficial.
It improves performance on perception-heavy questions but can degrade dialogue-centric reasoning, suggesting a visual distraction effect in speech-centered narrative video understanding.

Our main contributions are summarized as follows:
\begin{itemize}
\item \textbf{A diagnostic benchmark for the dialogue-to-plot evidence chain.}
We introduce CineMME, a human-verified benchmark that links temporal dialogue grounding, speech transcription, active-speaker localization, and plot-centric reasoning through shared short video clips.

\item \textbf{A densely annotated and human-verified narrative-video resource.}
CineMME contains $975$ curated short narrative video clips, $13,554$ speech segments, $33,083$ speaker face boxes, $1,600$ multiple-choice questions, and $316$ open-ended questions, and is constructed through automatic proposal generation, adversarial refinement, and human annotation and verification involving approximately $1,600$ expert person-hours.

\item \textbf{New diagnostics of model failure.}
Experiments with $20$ MLLMs on reasoning and $9$ omni-modal models on grounding reveal who-what-when decoupling, grounding-reasoning associations, and selective effects of visual input, which helps perception but can impair dialogue-centric reasoning.

\end{itemize}

\section{Related Works}
%\section{Related Works}
% \subsection{Multimodal Large Language Models}
\textbf{Multimodal Large Language Models. }MLLMs have progressed from image-language alignment~\cite{liuVisualInstructionTuning2023a, zhuMiniGPT4EnhancingVisionLanguage2023} to long-context video understanding.
Early Video-LLMs adapted visual encoders to LLMs through feature pooling~\cite{zhang2023video} or query-based alignment modules~\cite{li2023blip, dai2023instructblip}, while recent proprietary and open-weight systems expand context length, visual resolution, and instruction-following capabilities~\cite{OpenAI2025GPT5SystemCard, gemini-3-pro, zhang2024llavanextvideo, baiQwen3VLTechnicalReport2025}.
Omni-modal models further incorporate native audio streams, enabling models to process speech, sound, and visual events within a unified interface.
However, narrative video understanding stresses a capability beyond generic video description: models must bind dialogue, speaker identity, visual context, and narrative meaning over time.
CineMME evaluates this dialogue-to-plot evidence chain through temporal speech grounding, transcription, active-speaker localization, and downstream reasoning.

\textbf{Long-video and audio-visual QA.}
Video benchmarks have moved from short action-centric evaluation toward long-context and multimodal reasoning.
MVBench~\cite{liMVBenchComprehensiveMultimodal2024}, Video-MME~\cite{fuVideoMMEFirstEverComprehensive2025}, MLVU~\cite{zhouMLVUBenchmarkingMultitask2025}, and LongShOTBench~\cite{kurpathBenchmarkAgenticFramework2025} evaluate broad visual and temporal reasoning skills, while LongVALE~\cite{gengLongVALEVisionAudioLanguageEventBenchmark2025} and AV-Odyssey~\cite{gongAVOdysseyBenchCan2024} emphasize audio-visual understanding.
These benchmarks show the importance of multimodal input, but they generally evaluate final answer accuracy or event-level understanding rather than whether the relevant dialogue is localized, transcribed, and bound to the correct speaker.

\textbf{Movie understanding benchmarks.}
MovieQA~\cite{tapaswiMovieQAUnderstandingStories2016}, CinePile~\cite{rawalCinePileLongVideo2024}, and Movie101/Movie101v2~\cite{yueMovie101NewMovie2023,yueMovie101v2ImprovedMovie2025} directly target cinematic or movie-domain understanding.
They evaluate plot comprehension, narration, character interactions, and long-range story structure, making them close in domain to CineMME.
However, their supervision is primarily answer- or narration-level, and therefore does not directly diagnose whether a model grounds the speech evidence supporting an answer or distinguishes the active speaker from visually salient listeners.

\textbf{Grounding-centric evaluation.}
Grounding benchmarks such as VidSTG~\cite{zhangWhereDoesIt2020}, TimeLens~\cite{zhangTimeLensRethinkingVideo2025}, SVAG-Bench~\cite{hannanSVAGBenchLargeScaleBenchmark2025}, OmniGround~\cite{gaoOmniGroundComprehensiveSpatioTemporal2025}, and Video-MSR~\cite{zhuVideoMSRBenchmarkingMultihop2026} provide rigorous localization metrics for temporal boundaries, spatio-temporal action grounding, object-centric grounding, or multi-hop spatial reasoning.
However, most grounding tasks are object- or action-centric and are not designed for dialogue grounding in narrative videos, where the active speaker may be off-screen, partially visible, visually less salient than a listener, or embedded in a multi-person shot.
CineMME differs by centering grounding on dialogue: each prediction must account for when speech occurs, what is said, and which speaker face corresponds to the active speaker.

\begin{table*}[t]
\centering
\scriptsize
\caption{
Closest benchmark comparison. We keep the same comparison dimensions as the full benchmark table: dataset scale, annotation type, modality, reasoning formats, and free-form perception capabilities.
The main paper reports representative benchmarks from four closely related categories; a broader comparison is provided in Appendix~\ref{apptab:dataset_comparison}. 
``Anno.'' indicates annotation type (A: automatic, M: manual); in modality, ``A'' denotes audio, ``V'' video, and ``I'' image.}
\label{tab:dataset_comparison}
\setlength{\tabcolsep}{3.2pt}
\begin{tabular}{l|cccc|cc|ccc}
\toprule
& \multicolumn{4}{c|}{} & \multicolumn{2}{c|}{\textbf{Reasoning Q\&A}} & \multicolumn{3}{c}{\textbf{Free-Form Perception}} \\
\textbf{Benchmark} & \textbf{Videos} & \textbf{Q\&A} & \makecell{\textbf{Anno.}} & \textbf{Modality} &
\makecell{\textbf{MCQ}} &
\makecell{\textbf{Open}} &
\makecell{\textbf{Spatial-}\\\textbf{Grounding}} &
\makecell{\textbf{Temporal-}\\\textbf{Grounding}} & \textbf{Speech} \\
\midrule
\multicolumn{10}{l}{\violet{\textit{\textbf{Visual Benchmarks}}}} \\
\rowcolor{gray!10}
Video-MME~\citep{fuVideoMMEFirstEverComprehensive2025} & 900 & 2.7K & M & V & \cmark & \xmark & \xmark & \xmark & \xmark \\
MLVU~\citep{zhouMLVUBenchmarkingMultitask2025} & 1,730 & 3.1K & A+M & V & \cmark & \cmark & \xmark & \xmark & \xmark \\
\midrule
\multicolumn{10}{l}{\violet{\textit{\textbf{Omni Benchmarks}}}} \\
\rowcolor{gray!10}
LongVALE~\citep{gengLongVALEVisionAudioLanguageEventBenchmark2025} & 8.4K & -- & A & V+A & \xmark & \xmark & \xmark & \cmark & \cmark \\
AV-Odyssey~\citep{gongAVOdysseyBenchCan2024} & -- & 4.6K & A+M & I+A & \cmark & \xmark & \xmark & \xmark & \xmark \\
\midrule
\multicolumn{10}{l}{\violet{\textit{\textbf{Plot Benchmarks}}}} \\
\rowcolor{gray!10}
CinePile~\citep{rawalCinePileLongVideo2024} & 9,396 & 304K & A+M & V & \cmark & \xmark & \xmark & \xmark & \xmark \\
Movie101/v2~\citep{yueMovie101NewMovie2023,yueMovie101v2ImprovedMovie2025} & -- & -- & M & V & \xmark & \cmark & \xmark & \xmark & \xmark \\
\midrule
\multicolumn{10}{l}{\violet{\textit{\textbf{Grounding Benchmarks}}}} \\
\rowcolor{gray!10}
TimeLens-Bench~\citep{zhangTimeLensRethinkingVideo2025} & 4,279 & 6.5K & M & V & \xmark & \xmark & \xmark & \cmark & \xmark \\
SVAG-Bench~\citep{hannanSVAGBenchLargeScaleBenchmark2025} & 688 & 19.6K & A+M & V & \xmark & \xmark & \cmark & \cmark & \xmark \\
\midrule
\rowcolor{gray!10}
\textbf{\bench~(ours)} & \textbf{975} & \textbf{1.9K} & \textbf{A+M} & \textbf{V+A} & \cmark & \cmark & \cmark & \cmark & \cmark \\
\bottomrule
\end{tabular}
\vspace{-1em}
\end{table*}

Table~\ref{tab:dataset_comparison} focuses on the closest alternatives across four groups: visual QA, omni-modal understanding, movie understanding, and grounding-centric evaluation.
Overall, existing benchmarks evaluate important parts of video understanding, but they rarely connect speech transcription, temporal dialogue localization, active-speaker face grounding, and plot-centric reasoning within the same resource.
CineMME fills this gap by enabling diagnosis of not only whether a model answers a plot question correctly, but also whether it acquires and binds the multimodal evidence needed for that answer.

\section{Building CineMME: Coupled Grounding and Reasoning Data}
\label{sec:construction}
\begin{figure*}[t]
    \centering
    
    \includegraphics[width=\textwidth]{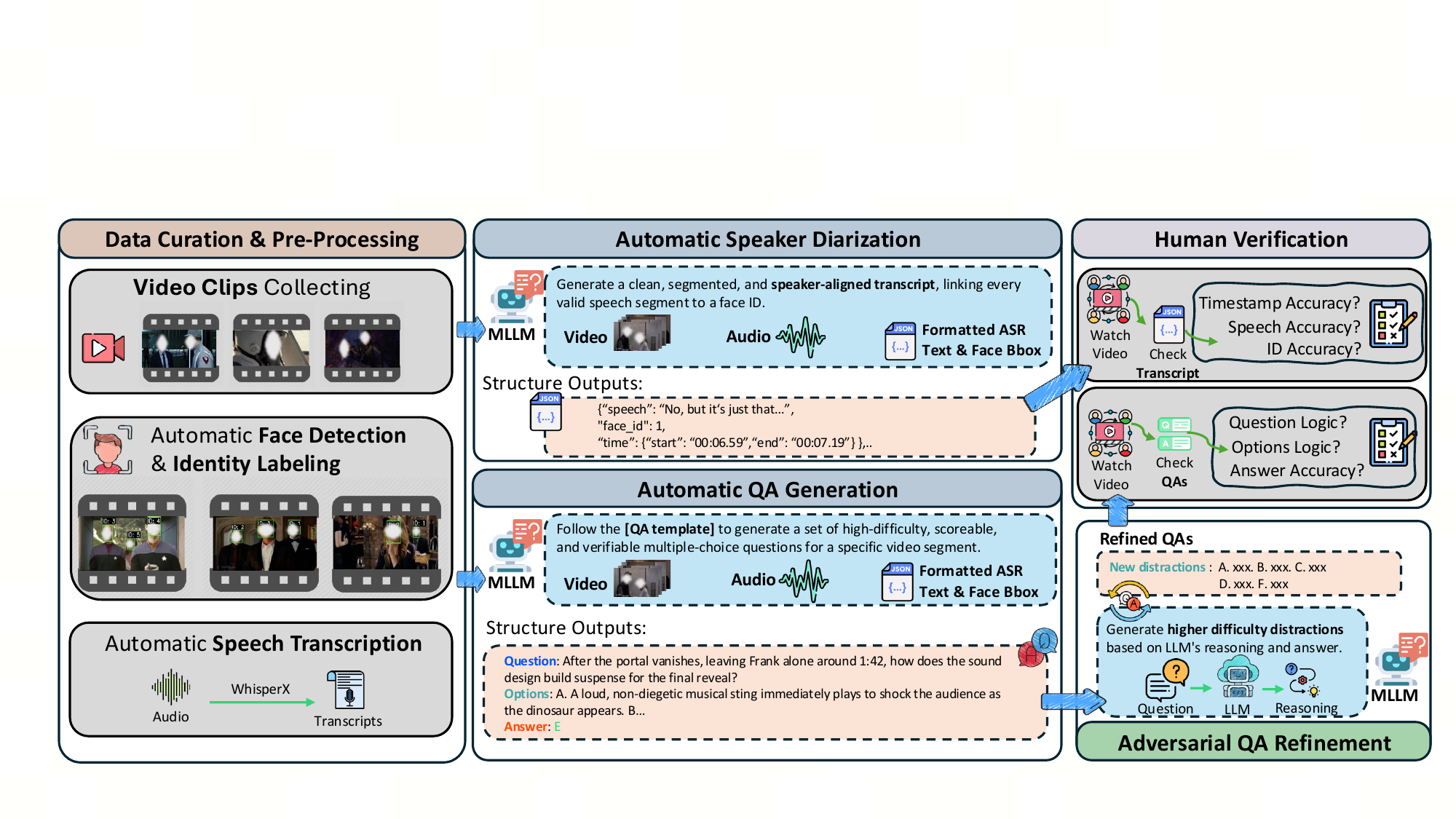}
    
    \caption{Annotation pipeline for \bench. We transform raw short narrative clips into a rigorous benchmark through a multi-stage process. Raw data undergoes preprocessing before entering parallel tracks for speaker diarization and QA generation. Crucially, we employ an \textit{Adversarial QA Refinement} strategy to enhance question difficulty, followed by a labor-intensive \textit{human verification} process to guarantee annotation quality.}
    \label{fig:overall_pipeline}
    \vspace{-2em}
\end{figure*}

CineMME is designed to support diagnostic evaluation of the dialogue-to-plot evidence chain, from dialogue grounding to speaker binding and downstream reasoning.
To balance annotation quality with scale, we construct the benchmark through a semi-automatic pipeline with three stages: \textit{clip curation and preprocessing}, \textit{adversarial question generation}, and \textit{human annotation and verification}.
Figure~\ref{fig:overall_pipeline} illustrates the overall framework.

\subsection{Data Curation and Preprocessing}
We curate $975$ high-quality short narrative video clips from publicly available online sources, covering diverse genres such as science fiction, drama, comedy, and action.
Clips are selected to capture self-contained narrative events with sufficient dialogue, character interaction, and narrative context.
Clip durations range from $22.1$ to $300.1$ seconds, with task-level medians of approximately $147\sim151$ seconds (Appendix~\ref{supp:clip_durations}).
This scope supports evaluation of evidence acquisition and reasoning within individual clips.

For each clip, we extract aligned speech and visual metadata.
On the audio side, we use WhisperX~\cite{bainWhisperXTimeAccurateSpeech2023} to obtain timestamped ASR transcripts.
On the visual side, we use InsightFace~\cite{deng2018arcface} for frame-level face detection and trajectory clustering, producing candidate participant track IDs.
The audio and visual streams are then synchronized to form candidate character-dialogue pairs, which serve as initial proposals for temporal speech localization, transcription, and active-speaker grounding.
Detailed preprocessing settings are provided in Appendix~\ref{supp:preprocessing}.

\subsection{Adversarial Question Generation}
Using the extracted speech and face metadata, we prompt proprietary MLLMs to generate initial speaker attribution labels and reasoning questions.
For CineMME-Reasoning, we target five categories: \textit{Perception}, \textit{Audio}, \textit{Narrative}, \textit{Social}, and \textit{Scene Presentation} understanding.
However, directly generated multiple-choice questions often contain shortcut cues, allowing models to answer based on language priors, salient visual hints, or implausible distractors.

To reduce this \textit{simplicity bias}, we apply adversarial QA refinement.
Lightweight open-weight Video-LLMs are first used as adversaries to answer candidate MCQs.
If an adversary answers a question correctly, the instance is flagged for refinement as potentially insufficiently discriminative.
We then provide the adversary's answer and reasoning trace to a stronger teacher model, which generates more plausible distractors targeting the apparent shortcut.
This refinement is intended to encourage questions that require grounded multimodal evidence rather than pattern matching.
Additional analysis of the refinement strategy is provided in Appendix~\ref{supp:adv_refine}.

\subsection{Human Annotation and Verification}
All final annotations are established through human annotation followed by independent review.
For each data point, an annotator checks and corrects the machine-generated proposals, and a second annotator reviews the resulting annotations.
The reviewer has access to the preceding annotations; the two stages form a sequential quality-control process.

For \textit{CineMME-Grounding}, annotators verify ASR text, timestamp boundaries, speaker identity, face-track IDs, and active-speaker boxes.
Timestamp boundaries are checked with a tolerance of less than $1$ second.
For \textit{CineMME-Reasoning}, annotators verify whether each question is answerable from the video and whether its reference answer is supported by the clip.
For MCQs, they additionally check that there is a unique correct option and that the distractors are plausible.
Invalid or ambiguous questions are discarded or, when suitable for open-ended evaluation, converted into open-ended questions.
Based on platform logs, the annotation effort is estimated at approximately $1{,}600$ expert person-hours: about $850$ hours for Grounding annotation, $600$ hours for Reasoning annotation, and $150$ hours for the final cross-track audit.
We compute Cohen's $\kappa$ on categorical verification decisions recorded in both the annotation and review stages, obtaining $0.89$--$0.94$ across ASR, timestamp, speaker track-ID, and discard decisions.
These values summarize consistency across sequential review stages.
The final verification logs mark $10.19\%$ of candidate speech segments for discarding.
The full rubric, effort breakdown, per-field sample sizes, and final verification outcomes are provided in Appendix~\ref{supp:human_eval}.

\subsection{Task Design: From Evidence Acquisition to Reasoning}
\label{sec:tasks}
CineMME is designed to evaluate the dialogue-to-plot evidence chain from low-level evidence acquisition to high-level narrative reasoning.
Accordingly, we define two linked tracks: \textit{CineMME-Grounding}, which evaluates whether models can acquire and bind dialogue-level evidence, and \textit{CineMME-Reasoning}, which evaluates whether models can use such evidence for narrative video understanding.

\paragraph{CineMME-Grounding: Dialogue-Level Evidence Acquisition.}
This track evaluates fine-grained audio-visual grounding from raw short narrative clips.
Unlike video captioning or holistic QA, this task requires the model to jointly predict three aligned outputs for each speech event:
1) the speech transcript,
2) the temporal boundaries, specified by start and end timestamps, and
3) the active speaker's location, represented by face bounding boxes.
This design directly tests whether models can answer the \emph{when}, \emph{what}, and \emph{who} components of the dialogue-to-plot evidence chain.

To balance evaluation precision and computational cost, we adopt a sparse spatial evaluation protocol.
Instead of requiring dense frame-by-frame tracking, models predict active-speaker face boxes at integer-second timestamps, \ie, at $1$ FPS, within each active speech segment.
This setting reduces token and annotation overhead while still measuring whether the model can maintain speaker-level spatial grounding over time.

We evaluate grounding with three complementary metrics.
1) Temporal IoU (tIoU) measures the alignment between predicted and ground-truth speech segments.
2) Word Error Rate (WER) measures transcription error on temporally matched segments.
3) Video IoU (vIoU) measures active-speaker localization over the union of predicted and annotated track durations.
Together, these metrics decompose dialogue grounding into temporal localization, speech transcription, and speaker binding.

Appendix~\ref{supp:metrics} provides the metric definitions and computation details.

\begin{figure*}[t]
   \centering
   \subcaptionbox{Segment Duration\label{fig:pie-segment}}[0.19\textwidth]{
    \includegraphics[width=\linewidth]{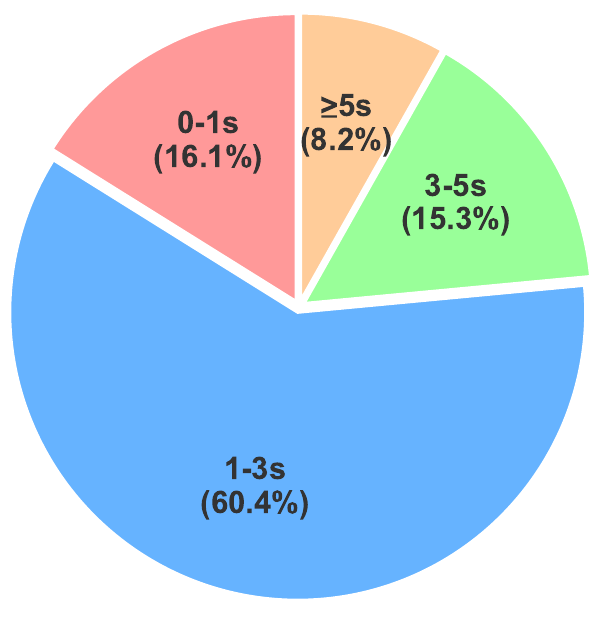}
   }
   \vspace{-0.1em}
   \hfill
   \subcaptionbox{Box Size\label{fig:pie-bbox}}[0.19\textwidth]{
    \includegraphics[width=\linewidth]{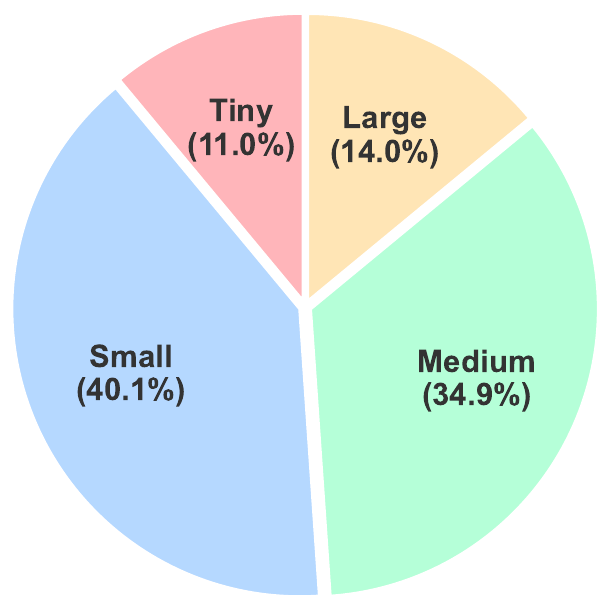}
   }
   \vspace{-0.1em}
   \hfill
   \subcaptionbox{Task Type\label{fig:pie-tasktype}}[0.19\textwidth]{
    \includegraphics[width=\linewidth]{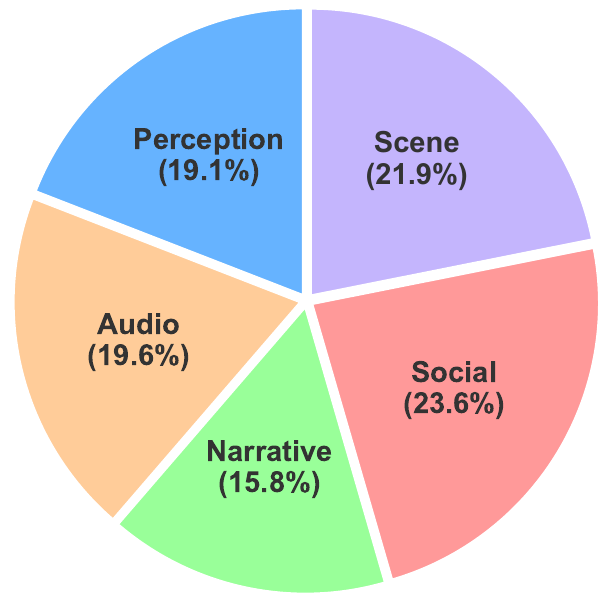}
   }
   \vspace{-0.1em}
   \hfill
   \subcaptionbox{Question Length\label{fig:pie-qlen}}[0.19\textwidth]{
    \includegraphics[width=\linewidth]{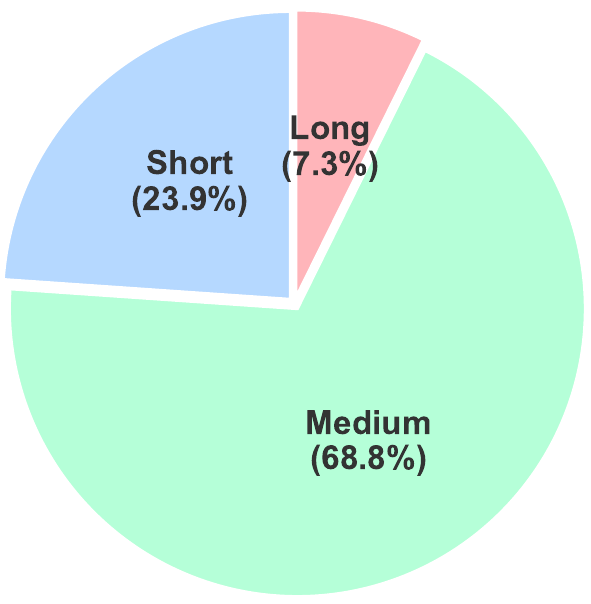}
   }
   
   \caption{
   Dataset statistics highlighting the diagnostic characteristics of \bench. 
   (a) Short speech segments require models to localize dialogue with high temporal precision. 
   (b) Tiny, small, and medium speaker face boxes make speaker-level grounding challenging, especially in multi-person narrative scenes.
   (c) Questions are balanced across five reasoning dimensions, supporting broad evaluation beyond a single capability. 
   (d) A large fraction of questions are medium or long, with $76.1\%$ exceeding $20$ words, encouraging semantic understanding beyond keyword matching.
   }
   \label{fig:dataset-stats}
   \vspace{-1em}
\end{figure*}

\paragraph{CineMME-Reasoning: Evidence-Grounded Plot Understanding.}
This track evaluates whether models can use dialogue, visual, and acoustic evidence to answer higher-level questions.
We organize questions into five categories: \textit{Perception}, \textit{Audio}, \textit{Narrative}, \textit{Social}, and \textit{Scene Presentation} understanding (Figure~\ref{fig:pie-tasktype}).
These categories cover complementary evidence needs, including objective visual facts, acoustic cues, story structure, character relationships, and visual composition.
All questions are human-verified for video dependence and support from the clip, with MCQs additionally checked for a unique correct option.
Detailed category descriptions and examples are provided in Appendix~\ref{supp:reasoning_taxonomy}.

We use two evaluation formats.
First, multiple-choice questions (MCQs) serve as the primary standardized evaluation, and we report accuracy overall and by category.
MCQs allow controlled comparison across models and support fine-grained analysis of capabilities such as audio reasoning, social inference, and scene-presentation understanding.
Second, open-ended questions (OEQs) evaluate generative reasoning beyond fixed options, especially for cases requiring multi-step explanation or narrative interpretation.
Because exact string matching is unsuitable for OEQs, we adopt a reference-based LLM-as-a-judge protocol.

For the main OEQ evaluation, we use Qwen2.5-72B-Instruct~\cite{qwen2.5} as an open-weight judge for reproducibility and cost efficiency.
The judge compares each model response with a human-verified reference answer and assigns a score from $0$ to $10$ based on reasoning alignment and factual consistency.
To validate the judge, we sample $110$ OEQ responses and obtain scores from two human experts.
The judge's scores correlate with the mean human scores at Pearson $r=0.758$, while the correlation between the two human experts is $0.752$.
These measurements provide supporting evidence of alignment with human scoring on this calibration sample.
The full scoring prompt and validation details are provided in Appendix~\ref{supp:judge_bias}.
% We employ \textbf{Multiple-Choice QA (MCQ)} for objective scoring across all dimensions. For open-ended generation, we implement a reference-based \textbf{LLM-as-a-Judge} evaluation. To ensure reproducibility and cost-effectiveness, we utilize \textbf{Qwen2.5-72B}~\citep{qwen2.5}, a high-performance open-weight model, as the judge. It scores responses against human-verified Gold Standards based on reasoning alignment and factual consistency. The scoring prompt and rubric are provided in Appendix~\ref{supp:judge_prompt}.

\subsection{Dataset Statistics and Diagnostic Characteristics}
\label{sec:stats}
CineMME contains $975$ unique short narrative video clips across two linked tracks, with $546$ clips for \textit{CineMME-Grounding} and $667$ clips for \textit{CineMME-Reasoning}.
The two tracks overlap on $238$ clips.
We summarize its diagnostic characteristics along three dimensions: fine-grained grounding, reasoning diversity, and linguistic characteristics.

\noindent\textbf{Fine-grained spatio-temporal grounding.}
CineMME emphasizes small, transient, speaker-level evidence rather than only prominent foreground objects.
As shown in Figure~\ref{fig:pie-segment} and Figure~\ref{fig:pie-bbox}, $86.0\%$ of speaker face boxes occupy less than $15\%$ of the frame area (including the Tiny, Small, and Medium bins), and $16.1\%$ of speech segments last less than $1$ second.
These characteristics make spatial precision and fine-grained temporal localization important, particularly in clips containing reaction shots and multi-person scenes.
Face-box area is computed as normalized width multiplied by normalized height, with statistics calculated over all $33,083$ stored boxes; duration statistics are calculated over all $13,554$ speech segments.
\begin{wrapfigure}{r}{0.47\textwidth} % {r}代表图片靠右排版，{0.5\textwidth}代表占据50%的页面宽度
  \centering
  \includegraphics[width=0.85\linewidth]{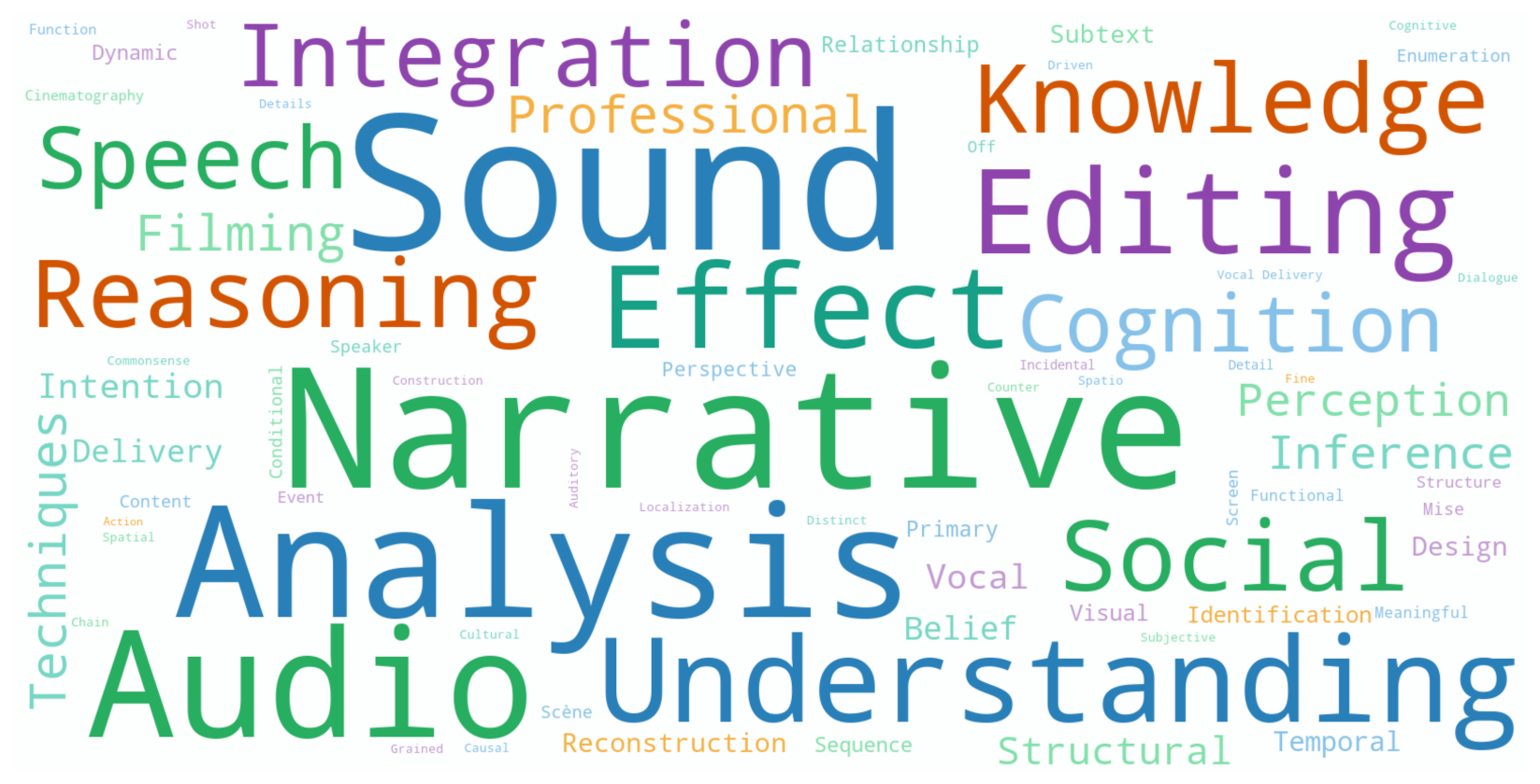}  
  \caption{Frequent question keywords center on reasoning, sound, editing, and narrative concepts.}
  \label{fig:wordcloud}
  \vspace{-1em}
\end{wrapfigure}

\noindent\textbf{Balanced narrative reasoning coverage.}
CineMME-Reasoning avoids concentrating on a single capability.
As shown in Figure~\ref{fig:pie-tasktype}, the $1,916$ QA pairs are distributed across five complementary categories: \textit{Perception} ($19.1\%$), \textit{Audio} ($19.6\%$), \textit{Narrative} ($15.8\%$), \textit{Social} ($23.6\%$), and \textit{Scene Presentation} ($21.9\%$).
This distribution supports a broad evaluation of how models use visual, acoustic, social, narrative, and scene-presentation cues.

\noindent\textbf{Linguistic and semantic complexity.} 
CineMME questions are designed to reduce reliance on shallow language priors.
As shown in Figure~\ref{fig:pie-qlen}, $76.1\%$ of questions contain more than $20$ words, and questions in the long-question bin contain more than $35$ words.
The word cloud in Figure~\ref{fig:wordcloud} summarizes recurring terms in the questions.
Keywords such as \textit{Reasoning}, \textit{Analysis}, and \textit{Understanding}, together with domain-specific terms such as \textit{Sound}, \textit{Editing}, and \textit{Narrative}, reflect the benchmark's thematic focus.
Detailed per-category word-count statistics are provided in Appendix~\ref{supp:stats}.

\noindent\textbf{Source availability.}
The July 2026 source audit retrieved metadata for $925$ of $975$ video IDs.
Restricting the dataset to these sources retains $98.44\%$ of MCQs, $95.57\%$ of OEQs, and $93.96\%$ of Grounding videos.
Appendix~\ref{supp:data_access} reports the retained subsets by category.
\section{Experiments}
\label{sec:experiments}
\begin{table*}[htbp]
  \centering
  \caption{Main Results on CineMME (Reasoning Track). We report MCQ accuracy (\%) and OEQ scores (0--100). \textbf{FPV}: Frames Per Video; \textbf{FPS}: Frames Per Second. Best scores within each modality/model group are in \textbf{bold}. MiMo-v2.5 is evaluated via API. Scores use valid answers; effective sample counts are in Table~\ref{tab:evaluation_coverage}.}
  \resizebox{\linewidth}{!}{
  \label{tab:main_results}
    \begin{tabular}{l c c | c c c c c c | c}
      \toprule
      \multirow{2}{*}{\textbf{Models}} & \multirow{2}{*}{\makecell{\textbf{LLM} \\ \textbf{Params}}} & \multirow{2}{*}{\makecell{\textbf{Frame} \\ \textbf{Rate}}} & \multicolumn{6}{c|}{\textbf{MCQs Accuracy (\%)}} & \multirow{2}{*}{\textbf{OEQs}} \\
      \cmidrule(lr){4-9}
       & & & Narrative & Percept. & Scene & Social & Audio & \textbf{Avg.} & \\
      \midrule
      \tabtitlecolor
      \multicolumn{10}{c}{\textit{\textbf{Open-source / open-weight MLLMs}}} \\
      \midrule
      \multicolumn{10}{l}{\textit{\textbf{Vision-Only}}} \\
      \midrule
      Phi-4-multimodal~\cite{microsoftPhi4MiniTechnicalReport2025} & 5.6B & 64 FPV & 17.65 & 20.15 & 28.74 & 24.34 & 27.92 & 24.22 & 28.63 \\
      Kimi-VL~\cite{teamKimiVLTechnicalReport2025} & 16B & 128 FPV & 18.04 & 20.90 & 20.92 & 23.87 & 27.83 & 22.56 & 37.82 \\
      InternVL3.5~\cite{wangInternVL35AdvancingOpenSource2025} & 38B & 32 FPV & 27.45 & 28.73 & 36.39 & 27.68 & 33.66 & 30.88 & 29.46 \\
      Ministral-3~\cite{liuMinistral32026} & 14B & 64 FPV & 18.04 & 23.13 & 24.36 & 20.05 & 27.83 & 22.69 & 30.51 \\
      MiMo-VL~\cite{liXiaomiMiMoVLMilocoTechnical2025} & 7B & 1 FPS & 26.67 & 25.00 & 33.81 & 23.15 & 31.07 & 27.88 & 23.43 \\
      GLM-4.1V~\cite{teamGLM45VGLM41VThinkingVersatile2025} & 9B & 1 FPS & 22.35 & 21.27 & 26.07 & 21.96 & 24.27 & 23.25 & 20.89 \\
      GLM-4.6V-Flash~\cite{teamGLM45VGLM41VThinkingVersatile2025} & 9B & 1 FPS & 27.15 & 30.70 & 36.81 & 28.41 & 32.21 & 31.11 & 29.74 \\
      GLM-4.6V~\cite{teamGLM45VGLM41VThinkingVersatile2025} & 100B & 1 FPS & 30.59 & 30.97 & 33.81 & 28.64 & 39.16 & 32.50 & \textbf{41.14} \\
      Qwen3-VL (8B)~\cite{baiQwen3VLTechnicalReport2025} & 8B & 1 FPS & 21.96 & \textbf{34.33} & 31.23 & 30.79 & 29.13 & 29.75 & 27.56 \\
      Qwen3-VL (32B)~\cite{baiQwen3VLTechnicalReport2025} & 32B & 1 FPS & 26.27 & 31.34 & 38.40 & 29.12 & \textbf{40.13} & 33.19 & 36.90 \\
      Qwen2.5-VL~\cite{baiQwen25VLTechnicalReport2025} & 72B & 1 FPS & \textbf{31.37} & 32.84 & \textbf{38.97} & \textbf{39.38} & 39.16 & \textbf{36.88} & 31.84 \\
      \midrule
      \multicolumn{10}{l}{\textit{\textbf{Audio-Capable}}} \\
      \midrule
      MiniCPM-o-2.6~\cite{yaoMiniCPMVGPT4VLevel2024} & 8B & 1 FPS & 20.00 & 23.13 & 22.92 & 19.57 & 27.18 & 22.44 & 26.99 \\
      Omnivinci~\cite{yeOmniVinciEnhancingArchitecture2025} & 9B & 1 FPS & 22.35 & 34.08 & 32.66 & 25.54 & 33.22 & 29.49 & 28.52 \\
      Qwen2.5-Omni~\cite{xuQwen25OmniTechnicalReport2025} & 7B & 0.5 FPS & 21.26 & 20.97 & 22.92 & 25.36 & 28.90 & 24.12 & 17.66 \\
      Qwen3-Omni~\cite{xuQwen3OmniTechnicalReport2025} & 30B & 1 FPS & 30.98 & 47.57 & 37.54 & 37.47 & 38.51 & 38.34 & 29.62 \\
      MiniCPM-o-4.5~\cite{cui2026minicpmo45realtimefullduplex} & 9B & 128 FPV & 26.19 & 34.87 & 31.98 & 28.88 & 34.97 & 31.30 & 33.85 \\
      MiMo-v2.5~\cite{xiaomi2026mimo25} & 310B & 1 FPS & \textbf{60.78} & \textbf{66.79} & \textbf{50.14} & \textbf{55.61} & \textbf{62.46} & \textbf{58.44} & \textbf{42.25} \\
      \midrule
      \tabtitlecolor
      \multicolumn{10}{c}{\textit{\textbf{Proprietary MLLMs}}} \\
      \midrule
      \multicolumn{10}{l}{\textit{\textbf{Vision-Only}}} \\
      \midrule
      GPT-5~\cite{OpenAI2025GPT5SystemCard} & -- & 100 FPV & 60.78 & 64.18 & 64.47 & 63.48 & 64.08 & 63.50 & 37.41 \\
      \midrule
      \multicolumn{10}{l}{\textit{\textbf{Audio-Capable}}} \\
      \midrule
      Qwen3.5-Omni-Plus~\cite{qwenteam2026qwen35omnitechnicalreport} & -- & 1 FPS & 57.79 & 66.87 & 58.96 & 55.69 & \textbf{68.72} & 61.20 & 43.87 \\
      Gemini-3 Pro~\cite{gemini-3-pro} & -- & 1 FPS & \textbf{65.74} & \textbf{79.39} & \textbf{59.25} & \textbf{68.04} & 68.21 & \textbf{67.66} & \textbf{45.64} \\
      \bottomrule
    \end{tabular}}
\end{table*}

We evaluate MLLMs on \bench\ to diagnose dialogue-level grounding, clip-level reasoning, modality effects, and grounding--reasoning relationships.

\subsection{Experimental Setup}
\label{sec:setup}
\noindent\textbf{Models and inputs.}
The main evaluation compares 20 models on reasoning and nine audio-capable models on grounding.
Table~\ref{tab:main_results} lists the frame-sampling settings for each model.
MiniCPM-o-4.5 uses 128 uniformly sampled frames and full audio for MCQ, OEQ, and grounding.
Gemini-3 Pro receives native video through the API at the default 1 FPS for both tracks.
MiMo-v2.5 has publicly released weights and is evaluated via API.
Audio and video are supplied separately for vLLM runs; hosted models use their multimodal APIs.
Model configurations and prompts are provided in Appendix~\ref{supp:model_details}.

\noindent\textbf{Scoring.}
For the main results, MCQ accuracy is the proportion of correct answers among extractable responses.
OEQ scores average valid judge ratings on a 0--10 scale and are multiplied by ten.
Grounding scores are video macro averages over parsed predictions; vIoU excludes videos without annotated face boxes.
Effective sample counts are reported in Table~\ref{tab:evaluation_coverage}, and metric details are in Appendix~\ref{supp:metrics}.
API content-moderation failures (\texttt{DataInspectionFailed}) reduce Qwen3.5-Omni-Plus coverage; its MCQ and grounding scores are conditional on valid responses (Appendix~\ref{supp:stat_sig}).

\subsection{Main Results}
\label{sec:main_results}
\noindent\textbf{Reasoning.}
Gemini-3 Pro achieves the highest MCQ accuracy among the evaluated models at 67.66\%, followed by GPT-5 at 63.50\% and Qwen3.5-Omni-Plus at 61.20\%.
MiMo-v2.5 achieves the highest open-weight MCQ accuracy at 58.44\%, which is 9.22 percentage points below Gemini-3 Pro.
Appendix~\ref{supp:stat_sig} reports bootstrap confidence intervals.

CineMME complements general video QA benchmarks such as Video-MME~\cite{fuVideoMMEFirstEverComprehensive2025} and Video-MMMU~\cite{huVideoMMMUEvaluatingKnowledge2025} by focusing on dialogue-grounded plot understanding.
The category breakdown highlights distinct evidence needs: perception-heavy questions benefit from visual capacity, whereas narrative, social, and audio-centric questions require models to track dialogue, sound, character intent, and temporal context.

\begin{table}[t]
  \centering
  \caption{Zero-shot grounding results on \textbf{\tracka}. All models predict transcripts, timestamps, and active-speaker face boxes. Scores are percentages; higher tIoU/vIoU and lower WER are better. MiMo-v2.5 is open-weight and evaluated via API.}
  \label{tab:open_closed_llm_method}
  \resizebox{0.7\linewidth}{!}{
    \begin{tabular}{lcccc}
      \toprule
      \textbf{Model} & \textbf{Params} & \textbf{tIoU $\uparrow$} & \textbf{vIoU $\uparrow$} & \textbf{WER $\downarrow$} \\
      \midrule
      \tabtitlecolor \multicolumn{5}{c}{\textit{Open-source / open-weight omni-modal MLLMs}} \\
      \midrule
      MiniCPM-o-2.6~\cite{yaoMiniCPMVGPT4VLevel2024} & 8B & 34.32 & 0.24 & 96.76 \\
      OmniVinci~\cite{yeOmniVinciEnhancingArchitecture2025} & 9B & 5.13 & 0.09 & 98.91 \\
      Phi-4-multimodal~\cite{microsoftPhi4MiniTechnicalReport2025} & 5.6B & 16.96 & 0.09 & 98.22 \\
      Qwen2.5-Omni~\cite{xuQwen25OmniTechnicalReport2025} & 7B & 24.72 & 0.38 & 93.31 \\
      Qwen3-Omni~\cite{xuQwen3OmniTechnicalReport2025} & 30B & 49.25 & 3.18 & 75.79 \\
      MiniCPM-o-4.5~\cite{cui2026minicpmo45realtimefullduplex} & 9B & 7.76 & 0.12 & 99.49 \\
      MiMo-v2.5~\cite{xiaomi2026mimo25} & 310B & \textbf{65.33} & \textbf{4.90} & \textbf{38.12} \\
      \midrule
      \tabtitlecolor \multicolumn{5}{c}{\itshape Proprietary omni-modal MLLMs} \\
      \midrule
      Gemini-3 Pro~\cite{gemini-3-pro} & -- & 68.34 & \textbf{7.74} & 28.99 \\
      Qwen3.5-Omni-Plus~\cite{qwenteam2026qwen35omnitechnicalreport} & -- & \textbf{68.74} & 6.45 & \textbf{28.87} \\
      \bottomrule
    \end{tabular}
   }
   \vspace{-2ex}
\end{table}

\noindent\textbf{Grounding.}
Table~\ref{tab:open_closed_llm_method} evaluates nine omni-modal models on dialogue-level grounding.
The central pattern is a clear \emph{who-what-when decoupling}.
MiMo-v2.5 and Gemini-3 Pro partially recover when dialogue occurs, reaching 65.33\% and 68.34\% tIoU, with WERs of 38.12\% and 28.99\%, respectively.
However, they remain weak at grounding who is speaking: Gemini-3 Pro achieves the highest vIoU at only 7.74\%, while Qwen3.5-Omni-Plus reaches 6.45\% despite similar temporal and transcription scores (68.74\% tIoU and 28.87\% WER).
Among open-weight models, MiMo-v2.5 reaches the highest vIoU at 4.90\%.
These results show that models can partially localize and transcribe dialogue while failing to bind utterances to the correct visible face.
This speaker-binding failure is especially relevant to narrative videos with reaction shots, multi-person scenes, and off-screen speech.

\subsection{Diagnostic Analyses and Ablations}
\label{sec:analysis}
\begin{table}[t]
  \centering
  \caption{Audio-only vs.\ audio+video ablation on Gemini-3 Pro~\cite{gemini-3-pro} and modality and visual sampling ablations on Qwen3-Omni~\cite{xuQwen3OmniTechnicalReport2025}. Effective sample counts are in Table~\ref{tab:ablation_coverage}.}
  \label{tab:gemini_av_ablation}
  \resizebox{\linewidth}{!}{
  \begin{tabular}{ccccccc|ccc}
    \toprule
    \multirow{2}{*}{\textbf{Model}} & \multirow{2}{*}{\textbf{Input}} & \multicolumn{5}{c|}{\textbf{\trackb~MCQ Accuracy (\%)}} & \multicolumn{3}{c}{\textbf{\tracka}} \\
    \cmidrule(lr){3-7}\cmidrule(lr){8-10}
     & & Narrative& Perception & Scene & Social & Audio  & tIoU & vIoU & WER$\downarrow$ \\
    \midrule
    Gemini-3-Pro & A & \textbf{69.02} & 67.54 & \textbf{63.61} & \textbf{70.88} & \textbf{74.43} & \textbf{68.64} & 3.46 & \textbf{27.42} \\
    Gemini-3-Pro & A+V & 65.74 & \textbf{79.39} & 59.25 & 68.04 & 68.21 & 68.34 & \textbf{7.74} & 28.99 \\
    \midrule
    Qwen3-Omni & A & \textbf{35.29} & 41.79 & 33.81 & 36.52 & \textbf{42.72} & \textbf{63.58} & 1.92 & \textbf{36.77} \\
    Qwen3-Omni & A+V (1 FPS) & 30.98 & \textbf{47.57} & \textbf{37.54} & \textbf{37.47} & 38.51 & 49.25 & \textbf{3.18} & 75.79 \\
    \midrule
    Qwen3-Omni & A+V (256 FPV) & \textbf{34.90} & \textbf{44.57} & \textbf{36.10} & 35.56 & 41.75 & 50.40 & 3.01 & \textbf{71.52} \\
    Qwen3-Omni & A+V (64 FPV) & 32.55 & 44.03 & 34.38 & 35.56 & 41.10 & 50.48 & 3.67 & 75.92 \\
    Qwen3-Omni & A+V (16 FPV) & 33.33 & 43.28 & 34.38 & \textbf{36.04} & \textbf{44.34} & \textbf{54.66} & \textbf{3.82} & 72.20 \\
    \bottomrule
  \end{tabular}}
  \vspace{-14pt}
\end{table}

\noindent\textbf{Visual Input Provides Selective Benefits.}
Modality ablations in Table~\ref{tab:gemini_av_ablation} show task-dependent gains from vision.
For Qwen3-Omni, audio-only improves temporal grounding over audio+video (63.58\% vs.\ 49.25\% tIoU) and performs better on Narrative and Audio reasoning, whereas audio+video improves Perception (47.57\% vs.\ 41.79\%).
For Gemini-3 Pro, audio+video raises Perception from 67.54\% to 79.39\%, but reduces Audio, Social, Narrative, and Scene Presentation scores.
It also improves speaker localization (7.74\% vs.\ 3.46\% vIoU) at a small transcription cost (28.99\% vs.\ 27.42\% WER).
Thus, visual input helps perception and speaker localization, but can distract dialogue-centric reasoning when visually salient content is not task-relevant.

\noindent\textbf{More Frames Do Not Resolve Speaker Grounding.}
Increasing Qwen3-Omni's visual sampling from 16 to 256 frames does not consistently improve reasoning or grounding.
MCQ accuracy ranges from 37.31\% to 39.40\%, and vIoU remains within 3.01--3.82\%, while tIoU fluctuates rather than increasing monotonically.
This suggests that the grounding bottleneck extends beyond the number of visual tokens: models need to select dialogue-relevant frames and bind acoustic cues to the correct speaker.
Complete sampling results are provided in Table~\ref{tab:main_ablation}.

\begin{table}[t]
\centering
\caption{Grounding quality and open-ended reasoning on overlapping clips. Better and Worse denote the two halves ranked by each grounding metric. OEQ scores are on a 0--100 scale; grouping details are in Appendix~\ref{supp:diagnostic_details}.}
\label{tab:joint_grounding_reasoning}
\setlength{\tabcolsep}{4.5pt}
\begin{tabular}{llccc}
\toprule
\textbf{Model} & \textbf{Split metric} & \textbf{Better} & \textbf{Worse} & \textbf{Gap} \\
\midrule
\multirow{3}{*}{Gemini-3 Pro}
& tIoU $\uparrow$ & 51.55 & 44.71 & $+6.84$ \\
& vIoU $\uparrow$ & 47.68 & 49.86 & $-2.17$ \\
& WER $\downarrow$ & 52.43 & 43.43 & $+9.00$ \\
\midrule
\multirow{3}{*}{Qwen3-Omni}
& tIoU $\uparrow$ & 31.63 & 28.73 & $+2.91$ \\
& vIoU $\uparrow$ & 27.71 & 32.59 & $-4.88$ \\
& WER $\downarrow$ & 31.15 & 29.04 & $+2.12$ \\
\bottomrule
\end{tabular}
\vspace{-2ex}
\end{table}

\noindent\textbf{Grounding Quality and Open-ended Reasoning.}
We use clips shared by the two tracks to examine whether better evidence acquisition is associated with stronger open-ended reasoning.
For each model and grounding metric, we split overlapping videos at the middle rank of grounding quality and compare their mean OEQ scores.
As shown in Table~\ref{tab:joint_grounding_reasoning}, Gemini-3 Pro's better-tIoU and lower-WER groups score 6.84 and 9.00 points higher, respectively; the corresponding Qwen3-Omni gaps are 2.91 and 2.12 points.
The vIoU splits show the opposite pattern, with gaps of $-2.17$ and $-4.88$ points.
This diagnostic separates the relationship of temporal and transcription quality with reasoning from that of speaker localization.
Additional MCQ overlap results and grouping details are provided in Appendix~\ref{supp:diagnostic_details}.

\noindent\textbf{Human Reference and Evidence Controls.}
Six graduate volunteers answered 161 MCQs, achieving item-level accuracy of $77.64\%$ (95\% CI: $70.60$--$83.39\%$; Table~\ref{tab:human_performance}).
In supplementary MiMo-v2.5 controls, question/options only achieves $33.19\%$, while aligned A/V reaches $54.94\%$ and shuffled audio reduces it to $43.34\%$.
Thus, language priors contribute, but aligned media provide additional evidence.
Gemini-3 Pro OEQ controls show the same pattern without answer options.
On wrong-clip subsets, same-source clips outperform different-source clips by only $0.33$--$2.61$ MCQ points across three models, suggesting limited benefit from source identity alone.
Appendix~\ref{supp:evidence_controls} gives the protocols, full results, and evaluation populations.

\section{Discussion and Limitations}
\label{sec:discussion}
\noindent\textbf{Interpretation.}
The results show that dialogue localization, transcription, and speaker binding should be evaluated separately; reasoning accuracy alone does not establish correct grounding.
The shared-clip analyses describe associations rather than causal effects.

\noindent\textbf{Scope and reproducibility.}
CineMME covers short English-language narrative clips, with limited coverage of other languages, cultures, production styles, and longer narratives.
Public source material may occur in pretraining; language priors and source recall can contribute to scores.
Although the evaluated models receive no CineMME-specific training, a public test set remains susceptible to future adaptation.
Third-party source media remain subject to copyright and platform terms, and source deletion can affect reproducibility.
Broader source coverage and hidden or periodically refreshed splits are future directions.

\section{Conclusion}
\label{sec:conclusion}

We introduced \textbf{\bench}, a human-verified benchmark for diagnosing dialogue-grounded plot understanding.
By linking dialogue-level grounding with high-level plot reasoning, CineMME evaluates whether models can recover when dialogue occurs, what is said, who says it, and why it matters.
Our experiments show that current omni-modal models remain weak at this evidence chain: they can partially localize and transcribe dialogue, but still struggle to bind utterances to the correct speaker, revealing a who-what-when decoupling.
Modality ablations further show that visual input is selectively useful, helping perception-heavy questions while sometimes distracting dialogue-centric reasoning. 
These findings suggest that future models need stronger mechanisms for selecting dialogue-relevant evidence, associating speech with characters in multi-person or off-screen settings, and integrating grounded evidence into narrative reasoning. 
We hope that CineMME can provide a compact diagnostic testbed for measuring such progress.

\section*{Acknowledgments}
This work is supported by the New Generation Artificial Intelligence-National Science and Technology Major Project (Grant No. 2025ZD0123505) and the National Natural Science Foundation of China (Grant Nos. 62550062, 62425606, 62576342, 32341009) and the Beijing Natural Science Foundation (Grant Nos. L252145, 4252054).

\medskip
{\small
\bibliographystyle{plain}
\bibliography{example_paper}
}
%%%%%%%%%%%%%%%%%%%%%%%%%%%%%%%%%%%%%%%%%%%%%%%%%%%%%%%%%%%%
\newpage

\appendix
% \onecolumn

\section{Societal Impact}
\label{supp:societal_impact}

CineMME is intended as a research benchmark for measuring fine-grained multimodal grounding and reasoning in short narrative video clips.
Its societal risks mainly concern responsible data access, speaker face-box annotations, cultural coverage, and possible misuse of automated narrative-understanding systems.

\noindent\textbf{Responsible release and use policy.}
The benchmark is built from publicly available online videos and is intended for non-commercial academic evaluation.
Benchmark resources include video identifiers, timestamps, annotations, prompts, evaluation code, and model outputs, together with retrieval and reconstruction instructions.
Source videos remain subject to the rights and terms of their respective owners and platforms.
Users are responsible for complying with applicable platform terms, institutional policies, and local requirements when accessing source media, and the benchmark should not be used to reconstruct source collections or support unauthorized redistribution.

\noindent\textbf{Face-box annotation risk.}
CineMME includes face detections and track IDs for speakers and participants in released metadata.
These annotations are used only for speaker grounding and plot understanding, not for identifying private individuals, biometric recognition, surveillance, or deepfake training.
We will state this restriction in the dataset license and documentation, and we discourage any attempt to merge these annotations with external identity databases.

\noindent\textbf{Cultural and domain bias.}
Because the source pool emphasizes publicly available English-language narrative videos, CineMME may overrepresent particular language communities, production styles, social norms, accents, and stereotypes.
Benchmark performance should therefore not be interpreted as a universal measure of video understanding across cultures or languages.
Future releases should expand multilingual, independent, documentary-style, and regionally diverse sources to reduce this bias.

\noindent\textbf{Misuse prevention and maintenance.}
The benchmark could be misused to optimize automated narrative analysis systems that support harmful surveillance, manipulative media generation, or unauthorized content retrieval.
We position CineMME strictly as an evaluation resource and will maintain a leaderboard with model/version metadata, modality settings, raw-output disclosure, and structured-output failure cases so that reported improvements remain auditable.

\section{Data Preprocessing Details}
\label{supp:implementation}

\subsection{Data Preprocessing Pipeline}
\label{supp:preprocessing}

To ensure high-quality multimodal alignment, we implement a rigorous preprocessing pipeline for both audio and visual streams. The specific algorithms and hyperparameters are detailed below.

\noindent\textbf{Audio Transcription and Alignment.}
We utilize the \textbf{WhisperX}~\citep{bainWhisperXTimeAccurateSpeech2023} framework to extract highly precise speech metadata. Specifically, we employ the \texttt{large-v3} model architecture for the initial Automatic Speech Recognition (ASR). To address the temporal imprecision inherent in raw ASR outputs, we subsequently apply a Forced Alignment step using a language-specific alignment model. This process yields word-level timestamps, ensuring that the boundaries of speech segments are synchronized with the visual stream.

\noindent\textbf{Visual Face Tracking and ID Assignment.}
For character grounding, we implement a custom online tracking algorithm leveraging \textbf{InsightFace}~\citep{deng2018arcface} for detection and feature extraction. The pipeline processes video frame-by-frame with the following logic:

\begin{enumerate}
    \item \textbf{Embedding Extraction:} For each detected face, we extract a facial embedding vector $v \in \mathbb{R}^{d}$ and apply L2 normalization: $\hat{v} = v / \|v\|_2$.
    \item \textbf{Centroid-based Assignment:} To ensure robust identification against pose variations, we maintain a history buffer $H_k = \{e_1, ..., e_t\}$ for each existing character ID $k$, storing the embeddings of the last $N=10$ frames. The identity of a new face $\hat{v}$ is determined by calculating the Cosine Distance to the centroid (mean embedding) of each existing cluster:
    \[
    d(\hat{v}, k) = 1 - \cos(\hat{v}, \frac{1}{|H_k|} \sum_{e \in H_k} e).
    \]
    \item \textbf{Update Rule:} The face is assigned to the ID with the minimum distance, provided the distance is below a threshold $\tau=0.7$. Upon assignment, the new embedding is added to the history buffer $H_k$ to update the centroid dynamically. If the minimum distance exceeds $\tau$, a new character ID is registered.
\end{enumerate}
This centroid-based approach significantly reduces ID switching caused by single-frame noise compared to simple frame-to-frame matching.

\subsection{Effectiveness of Adversarial Refinement}
\label{supp:adv_refine}

Standard automated QA generation often suffers from the issue of \textit{distractor triviality}, where automatically generated incorrect options lack plausibility or semantic relevance. Consequently, MLLMs can easily identify the correct answer through simple elimination strategies or linguistic shortcuts, rather than by relying on genuine video understanding.

To describe the effect of adversarial refinement, we track the performance of three representative models Qwen3-VL~\citep{baiQwen3VLTechnicalReport2025}(Qwen3-VL-30B-A3B-Instruct), Qwen2.5-VL~\citep{baiQwen25VLTechnicalReport2025} (Qwen2.5-VL-32B-Instruct) and GLM-4.1V~\citep{teamGLM45VGLM41VThinkingVersatile2025} (GLM-4.1V-9B-Thinking) across different optimization rounds. As shown in Table~\ref{tab:refine_appendix}, the initial generation yields higher accuracy scores (e.g., Qwen3-VL at 71.24\%). After two rounds of adversarial refinement, accuracy decreases across all three models (e.g., Qwen3-VL drops to 24.89\%). These results show that the refined questions are more challenging for the evaluated models.
The refinement prompt keeps the question and correct answer fixed and instructs the teacher to modify only distractors, which must remain incorrect and non-overlapping with the correct answer. The resulting MCQs undergo human annotation and review to check answer uniqueness and support from the video.

\begin{table}[h]
\centering
\caption{Adversarial refinement results. We report the zero-shot accuracy of three representative MLLMs across iterative optimization rounds. The decrease in accuracy summarizes the change in task difficulty for these models.}
\label{tab:refine_appendix}
\setlength{\tabcolsep}{6pt}
\begin{tabular}{l|ccc}
\toprule
Adv. Round &  Qwen2.5-VL  &  Qwen3-VL &  GLM-4.1V  \\
\hline
Initial & 68.24 & 71.24 & 57.08 \\
Round 1 & 46.78 & 44.64 & 35.19 \\
%\hline
\textbf{Round 2} & \textbf{20.17} & \textbf{24.89} & \textbf{11.16} \\
\bottomrule
\end{tabular}
\end{table}

\subsection{Human Evaluation Rubric}
\label{supp:human_eval}
Our human annotators were provided with a strict checklist to validate the dataset quality:
\begin{enumerate}
    \item \textbf{Temporal Precision:} Does the start/end timestamp capture the complete speech segment without cutting off words? (Tolerance: $\pm 1s$).
    \item \textbf{Visual Alignment:} Does the bounding-box ID correctly track the speaker's face?
    \item \textbf{Logical Soundness (QA):} Do the question and options make sense in the clip context?
    \item \textbf{Accuracy (QA):} Does the marked correct option accurately answer the question, and is it the only correct option?
\end{enumerate}
Machine-generated candidates are first checked and corrected by a human annotator, then examined by a separate reviewer who resolves remaining errors or uncertain cases.
Independent review refers to review by a different person with access to the preceding annotations.
Platform logs estimate approximately $1{,}600$ expert person-hours across human annotation, subsequent review, and final quality control.
The effort breakdown is reported in Table~\ref{tab:annotation_effort}, and agreement statistics between annotation and review stages are reported in Table~\ref{tab:iaa_appendix}.

\begin{table}[h]
\centering
\caption{Annotation effort estimated from platform logs. Track averages are computed over videos in each track; the total average is computed over unique clips.}
\label{tab:annotation_effort}
\setlength{\tabcolsep}{6pt}
\begin{tabular}{lccc}
\toprule
\textbf{Stage} & \textbf{Scope} & \textbf{Avg. time / unit} & \textbf{Person-hours} \\
\midrule
Grounding & 546 videos & $\sim$1.56 h & $\sim$850 \\
Reasoning & 667 videos & $\sim$0.90 h & $\sim$600 \\
Final audit & 975 clips & $\sim$0.15 h & $\sim$150 \\
\midrule
Total & 975 clips & $\sim$1.64 h & $\sim$1,600 \\
\bottomrule
\end{tabular}
\end{table}

\paragraph{Verification-decision agreement.}
The archived annotation logs contain $23,603$ candidate speech segments across $1,178$ annotation tasks.
We match the \texttt{Label} and \texttt{Check} stages by task ID and segment position within the same candidate list.
For each field, we exclude pairs with a missing decision in either stage.
ASR, timestamp, and discard decisions use \texttt{Yes}/\texttt{No}; track-ID decisions additionally retain \texttt{Not Sure} as a separate category.
We compute $\kappa=(p_o-p_e)/(1-p_e)$ on these categorical decisions, where $p_o$ is observed agreement and $p_e$ is expected agreement from the two stages' category proportions.
Table~\ref{tab:iaa_appendix} reports consistency across the sequential annotation and review stages.

\begin{table}[h]
\centering
\small
\caption{Verification-decision agreement between annotation and review. $N$ counts paired, non-missing decisions for each field. Final flag rates use all $23,603$ candidate speech segments in the final logs, as detailed below.}
\label{tab:iaa_appendix}
\setlength{\tabcolsep}{5pt}
\begin{tabular}{lrccc}
\toprule
\textbf{Item} & \textbf{$N$} & \textbf{Agreement} & \textbf{Cohen's $\kappa$} & \textbf{Final flag rate} \\
\midrule
ASR text & 21,771 & 95.01\% & 0.8998 & 51.52\% non-Yes \\
Timestamp & 21,833 & 94.56\% & 0.8910 & 55.25\% non-Yes \\
Track ID & 21,742 & 97.69\% & 0.9415 & 32.18\% non-Yes \\
Discard decision & 23,603 & 98.10\% & 0.8895 & 10.19\% discard \\
\bottomrule
\end{tabular}
\end{table}

\paragraph{Final verification outcomes.}
The last column uses all $23,603$ candidate speech segments as its denominator.
For ASR, timestamps, and track IDs, \emph{non-Yes} includes \texttt{No}, missing flags, and \texttt{Not Sure} when present.
The respective counts are $12,160$, $13,040$, and $7,595$, including $1,757$, $1,705$, and $1,784$ missing flags; the track-ID count also includes $185$ \texttt{Not Sure} flags.
These rates summarize verification outcomes and do not measure actual metadata edits.
The discard rate is $2,404/23,603=10.19\%$ of candidate speech segments.

\section{Evaluation Metrics and Protocols}
\label{supp:metrics}

\subsection{Grounding Track Metrics}

\noindent\textbf{Temporal IoU (tIoU).}
We calculate the Intersection over Union between the predicted segment $s^{p}$ and ground-truth segment $s^{gt}$. A prediction is considered a True Positive if $\text{tIoU} \geq 0.5$. Within each video, the metric is averaged across greedily matched pairs; the headline score then averages videos with parsed predictions:
\[
\mathrm{tIoU} = \frac{1}{N_{\mathrm{match}}} \sum_{i=1}^{N_{\mathrm{match}}} \frac{\mathrm{intersection}(s_i^{\mathrm{gt}}, s_i^{\mathrm{p}})}{\mathrm{union}(s_i^{\mathrm{gt}}, s_i^{\mathrm{p}})}
\]

\noindent\textbf{Video IoU (vIoU).}
We construct piecewise-constant face tracks from all box timestamps, independently of speech-segment matching.
Each timestamp $t_i$ represents $[t_i-0.5,t_i+0.5]$ seconds; overlapping neighboring cells are divided at their midpoint and clipped to the video boundaries.
Let $T_p,T_g$ denote the predicted and annotated temporal supports, and $B_p(t),B_g(t)$ their box sets at time $t$.
We compute
\begin{equation}
\mathrm{vIoU} = \frac{\displaystyle\int_{T_p\cap T_g} s\!\left(B_p(t),B_g(t)\right)\,\mathrm{d}t}{\left|T_p\cup T_g\right|},
\label{eq:viou}
\end{equation}
where $s$ is symmetric best-match spatial IoU. For nonempty box sets $P$ and $G$,
\[
s(P,G)=\frac{1}{2}\left(
\frac{1}{|P|}\sum_{p\in P}\max_{g\in G}\mathrm{IoU}(p,g)
+\frac{1}{|G|}\sum_{g\in G}\max_{p\in P}\mathrm{IoU}(p,g)
\right).
\]
The integral is computed over intersections of the time cells; times supported by only one track increase the union duration without contributing overlap.
Six videos without annotated face boxes are excluded, leaving 540 eligible videos.
The reported score is the mean video-level vIoU over eligible parsed predictions, expressed as a percentage; an empty predicted track receives zero.

\noindent\textbf{Word Error Rate (WER).}
WER is computed as the Levenshtein distance normalized by the sequence length, considering substitutions ($S$), deletions ($D$), and insertions ($I$):
\[
\mathrm{WER} = \frac{S + D + I}{N_{words}}
\]
The implementation lowercases matched text and clips per-video WER to $[0,1]$. A video with no matched segments receives WER $=1$.

\section{LLM-as-a-Judge Implementation}
\label{supp:judge_prompt}
We utilize \textbf{Qwen2.5-72B-Instruct} as the automated evaluator for the main OEQ results. This model was selected for its strong reasoning capabilities and open availability, ensuring the reproducibility of our metric. The judge evaluates the candidate answer against the Gold Standard (reference answer + reasoning trace) on a scale of 0-10.

\label{sec:appendix_eval_prompt}
To ensure consistent and rigorous scoring of open-ended responses, we utilize the following system prompt for our LLM judge. The prompt explicitly instructs the judge to penalize ``lucky guesses'' (correct answers derived from flawed reasoning) and prioritize reasoning alignment.

% Auto-generated by Python script
% Title: LLM-as-a-Judge

% Auto-generated by Python script
% Title: LLM-as-a-Judge
\begin{tcolorbox}[
    breakable,
    colback=gray!5, 
    colframe=gray!60!black,
    title=\textbf{LLM-as-a-Judge}, 
    fontupper=\small\ttfamily,
    boxrule=0.5mm,
    arc=2mm,
    bottom=2mm, top=2mm
]

You are an impartial and meticulous AI Judge specializing in evaluating Video Understanding and Reasoning models.

Your task is to evaluate a {[}Candidate Response{]} against a provided {[}Video Context{]} (Caption) and a {[}Gold Standard{]} (Reference Answer \& Reasoning Trace).

\#\#\# Evaluation Principles:

1. **Factuality based on Context**: The {[}Caption{]} is the ABSOLUTE GROUND TRUTH. If the candidate mentions visual/audio details that contradict the caption, it is a hallucination error.

2. **Reasoning Alignment**: The candidate has provided a step-by-step reasoning process (\textless{}Step 1\textgreater{}, \textless{}Step 2\textgreater{}...). You must verify if their logic aligns with the {[}Correct Reasoning Trace{]}. A correct answer derived from wrong reasoning (lucky guess) should receive a penalty.

3. **Completeness**: Does the candidate cover all key elements mentioned in the {[}Correct Answer{]}?

4. **Consistency**: The candidate's final \textless{}Answer\textgreater{} must be consistent with their own reasoning steps.

\#\#\# Scoring Scale (0-10):

- **10**: Perfect. Matches reference reasoning and answer completely.

- **8-9**: Correct answer, reasoning is good but misses minor nuance.

- **6-7**: Correct answer, but reasoning is vague or partially flawed.

- **4-5**: Answer is partially correct (captures part of the truth), or reasoning is weak or None.

- **2-3**: Incorrect answer, or significant hallucination in reasoning.

- **0-1**: Completely irrelevant or factually wrong.

\end{tcolorbox}

\begin{tcolorbox}[
    breakable,
    colback=gray!5, 
    colframe=gray!60!black,
    title=\textbf{Adversarial Distractor Generation Prompt}, % 建议将标题从 LLM-as-a-Judge 改为更贴合你评测 Benchmark 流程的名字
    fontupper=\small\ttfamily,
    boxrule=0.5mm,
    arc=2mm,
    bottom=2mm, top=2mm
]
\textbf{System Prompt:}\\
You are an expert AI system specialized in Adversarial AI Evaluation. Your objective is to generate high-quality, deceptive distractors (incorrect options) for multiple-choice questions to robustly evaluate model performance.

\textbf{[Input Context]}\\
You will be provided with items containing:\\
1. \textbf{Question \& Correct Answer:} Serves as the ground truth and stylistic anchor.\\
2. \textbf{Model Reasoning Trace (CoT):} The model's step-by-step logic demonstrating how it validates the correct answer and eliminates the original distractors.

\textbf{[Core Directives]}\\
\textbf{1. The Dual-Source Method}\\
- \textbf{Logic Source (via CoT Analysis):} Analyze the elimination logic within the CoT. If the model easily rejects original distractors for irrelevance or contradiction, formulate new distractors that are highly relevant and consistent with the premise. \textit{Objective:} Construct options that bypass initial rejection filters and target deep logical nuances (the "Pivot Point").\\
- \textbf{Style Source (via Correct Answer):} Ensure new distractors strictly mimic the correct answer in length, tone, grammatical structure, and complexity.

\textbf{2. Adversarial Strategies (Select based on CoT analysis)}\\
- \textbf{Fact/Task Mismatch:} The statement is contextually factual but addresses the wrong specific question.\\
- \textbf{Causal Reversal:} Inverts the cause-and-effect relationship.\\
- \textbf{Modifier Swap:} Introduces subtle changes to scope (e.g., substituting "Always" for "Sometimes", "Primary" for "Secondary").\\
- \textbf{Plausible Hallucination:} Utilizes correct domain terminology to describe a fabricated but realistic detail.\\
- \textbf{Misapplied Principle:} Applies a valid domain concept (referenced in the CoT) to an incorrect element.

\textbf{3. Safety Constraints}\\
- \textbf{Strictly Incorrect:} Distractors must be unequivocally incorrect. Discard any option with potential ambiguity.\\
- \textbf{Mutually Exclusive:} Distractors must not overlap in meaning with the Correct Answer.

\textbf{[Task Execution]}\\
For each provided question, improve ONLY the distractors using the principles above. Enhance their subtlety and deceptiveness without making them correct. 

\textbf{[Input Data]}\\
\{\{INPUT\_QUESTIONS\}\}

\textbf{[Output Specification]}\\
Return a valid JSON object strictly adhering to the schema below. Ensure the number of generated distractors exactly matches the \texttt{num\_distractors\_expected} parameter for each item. Do not include the correct answer in the distractor list. Do not include explanatory text.

\{ \\
\hspace*{2mm} "distractors": [ \\
\hspace*{4mm} \{ \\
\hspace*{6mm} "id": "<problem\_id\_1>", \\
\hspace*{6mm} "new\_distractors": ["<Option A>", "<Option B>", "<Option C>"] \\
\hspace*{4mm} \} \\
\hspace*{2mm} ] \\
\}
\end{tcolorbox}

\subsection{LLM Judge Validation}
\label{supp:judge_bias}

We validate \textbf{Qwen2.5-72B-Instruct} against human experts and a second LLM judge.

\begin{figure}[h]
    \centering
    \includegraphics[width=0.95\linewidth]{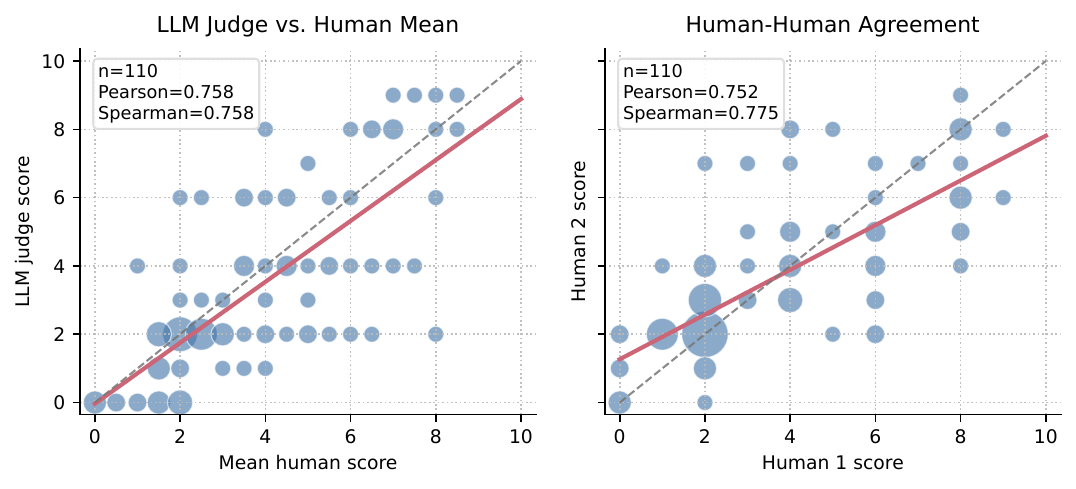}
    \caption{Human calibration of the OEQ judge. Left: LLM judge scores against the average of two human expert scores. Right: agreement between the two human experts. Marker size indicates repeated score pairs; the dashed diagonal shows exact agreement.}
    \label{fig:llm_judge_human_agreement}
\end{figure}

\noindent\textbf{Human calibration.}
We sample 110 OEQ responses from different evaluated models and ask two human experts to score them independently using the same 0--10 rubric as the LLM judge.
As shown in Table~\ref{tab:app_llm_judge_human_calibration} and Figure~\ref{fig:llm_judge_human_agreement}, the LLM judge correlates strongly with the mean human score (Pearson $r=0.758$, Spearman $\rho=0.758$, QWK $=0.721$).
The two human experts correlate at Pearson $r=0.752$, Spearman $\rho=0.775$, and QWK $=0.745$.
The judge-to-mean and human-to-human comparisons use different reference scores, so their magnitudes are reported descriptively. These results provide supporting evidence that the judge aligns with human scoring on this calibration sample.

\begin{table}[h]
\centering
\caption{Human calibration on 110 matched responses. Human mean averages two scores. QWK uses quadratic weights and rounds scores to the nearest integer.}
\label{tab:app_llm_judge_human_calibration}
\setlength{\tabcolsep}{4.5pt}
\begin{tabular}{lccccc}
\toprule
\textbf{Comparison} & \textbf{Pearson} & \textbf{Spearman} & \textbf{QWK} & \textbf{MAE} & \textbf{$\pm$1} \\
\midrule
Human 1 vs. Human 2 & 0.752 & 0.775 & 0.745 & 1.15 & 71.8\% \\
LLM judge vs. human mean & 0.758 & 0.758 & 0.721 & 1.28 & 58.2\% \\
LLM judge vs. Human 1 & 0.753 & 0.756 & 0.745 & 1.20 & 63.6\% \\
LLM judge vs. Human 2 & 0.659 & 0.673 & 0.638 & 1.55 & 51.8\% \\
\bottomrule
\end{tabular}
\end{table}

\begin{figure}[h]
    \centering
    \includegraphics[width=0.95\linewidth]{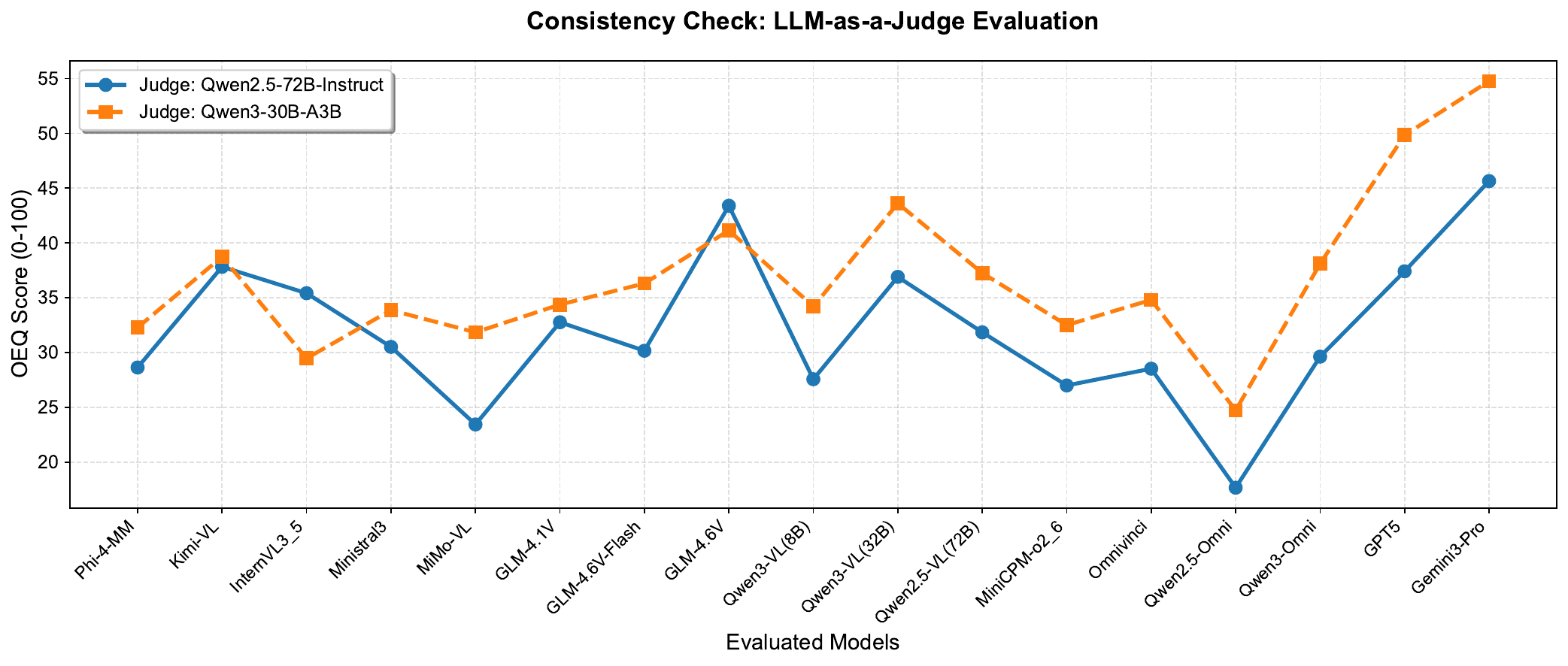}
    \caption{Consistency check of LLM-as-a-Judge. OEQ scores assigned by Qwen2.5-72B-Instruct and Qwen3-30B-A3B show strong agreement (Pearson $r=0.808$).}
    \label{fig:judge_compare}
\end{figure}

\noindent\textbf{Consistency Across Judges.}
We compare Qwen2.5-72B-Instruct with Qwen3-30B-A3B using the same textual references and scoring rubric.
Their OEQ scores correlate strongly (Pearson $r=0.808$; Figure~\ref{fig:judge_compare}), supporting the consistency of the model ranking across judges.

% \input{appendix_prompts/adv_refinement_prompt}

% \input{appendix_prompts/speaker_dia_prompt}

% \input{appendix_prompts/qa_task_prompt_v6}

% (Insert your prompt template here or leave as "See code for full prompt")

\section{CineMME-Reasoning Category Scope}
\label{supp:reasoning_taxonomy}

CineMME-Reasoning is organized around five high-level categories.
Table~\ref{tab:reasoning_taxonomy_appendix} summarizes the evaluation focus of each category and gives representative question coverage.
These descriptions clarify the scope of each category rather than define separate fine-grained evaluation tracks.

\begin{table*}[t]
\centering
\caption{Category scope of CineMME-Reasoning. The main text uses the five concise category names; this table clarifies the scope of each category with representative question coverage.}
\label{tab:reasoning_taxonomy_appendix}
\small
\setlength{\tabcolsep}{4pt}
\renewcommand{\arraystretch}{1.15}
\begin{tabular}{p{0.20\textwidth}p{0.36\textwidth}p{0.34\textwidth}}
\toprule
\textbf{Category} & \textbf{What It Measures} & \textbf{Representative Question Coverage} \\
\midrule
Perception & Extraction of objective audiovisual facts under visual complexity or short temporal windows. & Questions may ask about subtle visual details, dynamic counts, spoken content, speaker identity, or where and when an event occurs. \\
Audio & Interpretation of spoken and non-speech acoustic evidence, including subtle, masked, or off-screen cues. & Questions may focus on dialogue, vocal delivery, background sounds, subjective sound effects, or off-screen events that cannot be answered from frames alone. \\
Narrative & Reasoning about story structure, causal dependencies, chronology, and information asymmetry. & Questions may examine causal chains, temporal order, narrative structure, privileged information, or how one event changes the interpretation of another. \\
Social & Inference over intentions, beliefs, relationships, cultural context, and behavior that contradicts generic priors. & Questions may ask models to infer goals, false beliefs, implicit relationships, culturally grounded meanings, or evidence that overrides a generic social prior. \\
Scene Presentation & Understanding how visual composition and audio-visual presentation cues convey narrative, psychological, or thematic meaning. & Questions may connect framing, editing rhythm, sound design, lighting, props, or scene layout to their narrative or emotional effects. \\
\bottomrule
\end{tabular}
\end{table*}

\section{Human Performance and Evidence Controls}
\label{supp:evidence_controls}

We evaluate human performance, language priors, and the contribution of question-aligned media.
The supplementary MCQ comparisons use question-only and media-input baselines from the same control setup.
For these controls, MCQ accuracy is the fraction of correct answers among returned responses; unextractable answers receive zero, and failed API requests are excluded.
Overall accuracy weights individual questions equally.
The supplementary OEQ controls use NVIDIA-Nemotron-3-Nano-30B-A3B-BF16 with the reference-based 0--10 rubric, rescaled to 0--100; the main evaluation uses Qwen2.5-72B-Instruct.

\subsection{Human MCQ Performance}
\label{supp:human_performance}
Six graduate volunteers answered a subset of $161$ MCQs spanning all five categories.
Each question was independently answered by two volunteers using the clip-level evidence and without external source or transcript lookup.
Table~\ref{tab:human_performance} reports the item-level results: overall accuracy is $77.64\%$, with a Wilson 95\% confidence interval of $70.60$--$83.39\%$.
Perception reaches $89.19\%$, whereas Scene Presentation reaches $66.67\%$.
% Human scores: OpenReview author response RXIm0KA5jE. Gemini scores: tables/model_compare.tex.
\begin{table}[htbp]
\centering
\caption{Human MCQ performance on 161 questions answered by six volunteers. Gemini-3 Pro scores on the full benchmark are from Table~\ref{tab:main_results}. Differences are percentage-point gaps.}
\label{tab:human_performance}
\small
\setlength{\tabcolsep}{10pt}
\begin{tabular}{lrrr}
\toprule
Category & Human (\%) & Gemini-3 Pro (\%) & Difference \\
\midrule
Audio & 75.70 & 68.21 & +7.49 \\
Social & 78.26 & 68.04 & +10.22 \\
Narrative & 76.47 & 65.74 & +10.73 \\
Perception & 89.19 & 79.39 & +9.80 \\
Scene & 66.67 & 59.25 & +7.42 \\
\midrule
Overall & 77.64 & 67.66 & +9.98 \\
\bottomrule
\end{tabular}
\end{table}

\subsection{Modality Alignment and Language Priors}
\label{supp:modality_controls}
We compare five input conditions: question/options only, vision only, audio only, full audio--video, and target vision paired with an unrelated audio track.
For OEQ, the question-only condition contains no answer options.
The visual-only condition removes the original audio, while audio-only removes the visual input.
Shuffled audio preserves the target visual evidence but breaks its correspondence with the soundtrack.

Table~\ref{tab:modality_controls_mcq} gives the MiMo-v2.5 MCQ results.
Full A/V scores $54.94\%$, compared with $33.19\%$ for question/options only, $44.44\%$ for vision only, and $43.34\%$ with shuffled audio.
The $33.19\%$ text-only result exceeds the nominal six-option chance rate and reveals contributions from language priors.
Aligned media provide an additional gain: full A/V improves overall accuracy by $21.75$ points over text only and $11.60$ points over shuffled audio.
% Audit: nips26_re/camera_ready_mcq_audit_20261026/audit.json.
\begin{table}[htbp]
\centering
\caption{MiMo-v2.5 MCQ input controls. Accuracy (\%) is averaged over 1,600 scored responses per condition, except shuffled audio (1,599). Question/options only supplies no media; shuffled audio pairs the target video with an unrelated audio track.}
\label{tab:modality_controls_mcq}
\small
\setlength{\tabcolsep}{5pt}
\begin{tabular}{lrrrrrr}
\toprule
Input & Audio & Social & Narrative & Perception & Scene & Overall \\
\midrule
Question/options only & 39.16 & 32.46 & 31.76 & 30.22 & 32.09 & 33.19 \\
Vision only & 51.46 & 42.00 & 36.47 & 47.01 & 44.99 & 44.44 \\
Audio only & \textbf{57.93} & 47.97 & \textbf{48.63} & 58.21 & 42.69 & 50.56 \\
Full A/V & 57.28 & \textbf{53.94} & \textbf{48.63} & \textbf{64.18} & \textbf{51.58} & \textbf{54.94} \\
Vision + shuffled audio & 49.19 & 38.90 & 40.39 & 43.45 & 45.56 & 43.34 \\
\bottomrule
\end{tabular}
\end{table}

The Audio category covers sound-centered narrative reasoning, including vocal delivery, off-screen events, sound effects, and auditory perspective.
Some questions also admit partial inference from images.
Direct speech-content or speaker-identification questions can fall under Perception, while sound-design interpretation can fall under Scene Presentation.
On Audio, audio-only reaches $57.93\%$, full A/V $57.28\%$, vision-only $51.46\%$, and shuffled audio $49.19\%$.
Aligned audio improves accuracy on Audio questions over shuffled audio by $8.09$ points, while audio-only and full A/V perform similarly on this category.

The Gemini-3 Pro OEQ controls in Table~\ref{tab:modality_controls_oeq} provide evidence without answer options.
Full A/V scores $43.75$, versus $27.09$ for question only and $37.07$ with shuffled audio.
The improvement with aligned media therefore extends beyond choosing among MCQ distractors.
Table~\ref{tab:answer_positions} separately checks the balance of correct-option positions.
% Scores: final OpenReview author response RXIm0KA5jE.
% Model/judge identity: nips26_re/llm_judge_outputs/results/OEQ-05/06
% and llm_judge_outputs_extra/results/OEQMOD-07/08/09.
\begin{table}[htbp]
\centering
\caption{Gemini-3 Pro OEQ input controls, scored by NVIDIA-Nemotron-3-Nano-30B-A3B-BF16 with the reference-based rubric and rescaled to 0--100. $N$ counts valid judge scores out of 316 questions.}
\label{tab:modality_controls_oeq}
\small
\setlength{\tabcolsep}{4pt}
\begin{tabular}{lrrrrrrr}
\toprule
Input & Audio & Social & Narrative & Perception & Scene & Overall & \textit{N} \\
\midrule
Question only & 31.21 & 32.06 & 20.00 & 14.69 & 41.86 & 27.09 & 313 \\
Vision only & 31.36 & 46.47 & 33.33 & 26.22 & \textbf{58.57} & 38.33 & 311 \\
Audio only & 38.79 & 48.53 & \textbf{37.50} & 26.53 & 46.57 & 38.17 & 311 \\
Full A/V & \textbf{41.06} & \textbf{59.12} & 35.62 & \textbf{32.35} & 57.86 & \textbf{43.75} & 312 \\
Vision + shuffled audio & 34.09 & 38.24 & 34.79 & 28.78 & 51.43 & 37.07 & 314 \\
\bottomrule
\end{tabular}
\end{table}

\subsection{No-clip and Wrong-clip Controls}
\label{supp:movie_recall}
Table~\ref{tab:question_only_controls} compares question-only and correct-clip MCQ inputs for three models.
The full-input gains are $21.75$, $9.31$, and $18.38$ points for MiMo-v2.5, MiniCPM-o-4.5, and Qwen3-Omni, respectively.
All three show their largest category gain on Perception.
These controls quantify the contribution of local media beyond question text and possible source-level priors.
% Audit: nips26_re/camera_ready_mcq_audit_20261026/audit.json.
\begin{table}[htbp]
\centering
\caption{Question-only and correct-clip MCQ results from the supplementary control runs (1,600 scored responses per condition). Values are accuracy (\%); gains are percentage points computed before rounding.}
\label{tab:question_only_controls}
\small
\setlength{\tabcolsep}{4pt}
\begin{tabular}{llrrrrrr}
\toprule
Model & Input & Audio & Social & Narrative & Perception & Scene & Overall \\
\midrule
MiMo-v2.5 & Q/options only & 39.16 & 32.46 & 31.76 & 30.22 & 32.09 & 33.19 \\
 & Correct A/V & 57.28 & 53.94 & 48.63 & 64.18 & 51.58 & 54.94 \\
 & Gain & +18.12 & +21.48 & +16.86 & +33.96 & +19.48 & +21.75 \\
\midrule
MiniCPM-o-4.5 & Q/options only & 28.48 & 20.53 & 18.43 & 13.43 & 28.08 & 22.19 \\
 & Correct A/V & 33.33 & 27.92 & 31.37 & 37.69 & 29.51 & 31.50 \\
 & Gain & +4.85 & +7.40 & +12.94 & +24.25 & +1.43 & +9.31 \\
\midrule
Qwen3-Omni & Q/options only & 33.33 & 22.91 & 24.71 & 22.76 & 30.66 & 26.88 \\
 & Correct A/V & 44.98 & 47.26 & 38.82 & 50.00 & 44.13 & 45.25 \\
 & Gain & +11.65 & +24.34 & +14.12 & +27.24 & +13.47 & +18.38 \\
\bottomrule
\end{tabular}
\end{table}

For the wrong-clip analysis, we identify questions whose source narrative has another clip in the benchmark.
This yields $922$ MCQs and $170$ OEQs.
We keep each question and reference answer fixed and replace the media with another clip from the same source narrative or a clip from a different source narrative.
Both replacement conditions use the same eligible question sets.
The same-source condition retains source identity and associated prior knowledge while removing the target segment.
MiniCPM-o-4.5 and Qwen3-Omni receive audio and video; Qwen3-VL-30B-A3B receives vision.
% MCQ audit: nips26_re/camera_ready_mcq_audit_20261026/audit.json.
% OEQ sources: OpenReview responses RXIm0KA5jE / BJRVZeRIpr.
% Wrong-clip populations: nips26_re/wrong_clip_datasets/*_wrong_clip_summary.json.
% Correct/question-only rows are full-set references, not matched-subset estimates.
\begin{table}[htbp]
\centering
\caption{Source-recall controls. Q denotes question/options only; Correct is the target clip; Same and Different are wrong clips from the same and a different source narrative. Q and Correct use the full benchmark; the two wrong-clip conditions share 922 MCQs and 170 OEQs. OEQ uses the Nemotron judge specified in Table~\ref{tab:modality_controls_oeq}.}
\label{tab:wrong_clip_controls}
\small
\setlength{\tabcolsep}{4pt}
\begin{tabular}{llrrrrrrr}
\toprule
\multirow{2}{*}{Model} & \multirow{2}{*}{Media} & \multicolumn{4}{c}{MCQ accuracy (\%)} & \multicolumn{3}{c}{OEQ score (0--100)} \\
\cmidrule(lr){3-6}\cmidrule(lr){7-9}
 & & Q & Correct & Same & Different & Correct & Same & Different \\
\midrule
MiniCPM-o-4.5 & A/V & 22.19 & 31.50 & 22.78 & 22.45 & 28.79 & 23.31 & 18.76 \\
Qwen3-Omni & A/V & 26.88 & 45.25 & 32.00 & 29.39 & 28.15 & 16.12 & 13.14 \\
Qwen3-VL-30B-A3B & Vision & 28.00 & 35.62 & 32.97 & 30.37 & 30.35 & 21.01 & 19.76 \\
\bottomrule
\end{tabular}
\end{table}

Table~\ref{tab:wrong_clip_controls} reports the two wrong-clip conditions alongside the full-set question-only and correct-clip references.
For MiniCPM-o-4.5, Qwen3-Omni, and Qwen3-VL, respectively, the same-source advantage over different-source clips is $0.33$, $2.61$, and $2.60$ MCQ points, and $4.55$, $2.98$, and $1.25$ OEQ points.
The modest same-source advantage suggests that source identity provides limited assistance under these controls.
Together with the no-clip controls, these findings support the contribution of local evidence.

\section{Additional Statistics}
\label{supp:stats}
Questions average $25.6$ words, MCQ options $23.2$ words, and gold answers $17.5$ words.
We count whitespace-separated words without the A--F option labels; the overall means weight individual records or options equally.
\begin{table}[h]
\centering
\caption{Mean whitespace-separated word counts, excluding option labels. Questions and gold answers cover all 1,916 MCQ/OEQ records; options cover all 9,600 MCQ options. Overall means weight individual records or options equally.}
\label{tab:word_counts_appendix}
\begin{tabular}{l|ccc}
\toprule
Task Type & Question & Option & Answer \\
\midrule
Narrative & 25.6 & 23.7 & 19.0 \\
Perception & 24.5 & 13.2 & 10.1 \\
Scene & 26.6 & 28.3 & 22.3 \\
Social & 24.1 & 23.8 & 16.4 \\
Audio & 27.3 & 25.0 & 19.2 \\
\midrule
Overall & 25.6 & 23.2 & 17.5 \\
\bottomrule
\end{tabular}
\end{table}

\subsection{Clip Duration and Temporal Scope}
\label{supp:clip_durations}
Table~\ref{tab:clip_durations} reports the duration of the clip associated with each evaluation item.
MCQ and OEQ statistics weight questions equally; Grounding statistics weight videos equally.
The pooled $2{,}462$ observations include repeated clips across questions and tasks.
These short clips support dense alignment of speech, speakers, and local narrative evidence.
% Source: NeurIPS rebuttal BJRVZeRIpr; verified against
% nips26_re/analysis/results/clip_duration_stats/clip_duration_summary.json.
\begin{table}[htbp]
\centering
\caption{Clip durations in seconds. Each MCQ or OEQ contributes one observation, so clips with multiple questions occur repeatedly. Grounding contributes one observation per video. The pooled row combines task records rather than 975 unique clips. P90 is the 90th percentile.}
\label{tab:clip_durations}
\small
\setlength{\tabcolsep}{8pt}
\begin{tabular}{lrrrrrr}
\toprule
Track & \textit{N} & Min & Median & Mean & P90 & Max \\
\midrule
MCQ & 1600 & 52.1 & 151.1 & 156.0 & 205.2 & 299.1 \\
OEQ & 316 & 22.1 & 148.1 & 148.0 & 190.2 & 274.1 \\
Grounding & 546 & 22.1 & 147.1 & 148.0 & 189.1 & 300.1 \\
\midrule
Pooled task records & 2462 & 22.1 & 151.1 & 153.2 & 202.1 & 300.1 \\
\bottomrule
\end{tabular}
\end{table}

\subsection{Answer-Position Distribution}
\label{supp:answer_positions}
Table~\ref{tab:answer_positions} reports the distribution of correct MCQ options.
The shares range from $14.63\%$ to $18.50\%$, with no significant departure from uniform positions in a chi-square test ($p=0.125$).
% Source: NeurIPS rebuttal RXIm0KA5jE; counts verified against
% nips26_re/analysis/results/mcq_option_bias/mcq_option_bias_summary.json.
\begin{table}[htbp]
\centering
\caption{Correct-answer positions in the 1,600-question MCQ set. A chi-square goodness-of-fit test against uniform positions gives $\chi^2(5)=8.623$, $p=0.125$.}
\label{tab:answer_positions}
\small
\setlength{\tabcolsep}{12pt}
\begin{tabular}{lrrrrrr}
\toprule
Position & A & B & C & D & E & F \\
\midrule
Count & 275 & 252 & 267 & 296 & 276 & 234 \\
Share (\%) & 17.19 & 15.75 & 16.69 & 18.50 & 17.25 & 14.63 \\
\bottomrule
\end{tabular}
\end{table}

\subsection{Source Availability and Research Access}
\label{supp:data_access}
In July 2026, we audited all $975$ source video IDs through YouTube metadata requests: $925$ returned metadata and $50$ were classified as unavailable.
This measures source discoverability at the audit date; playback and retrieval may additionally depend on region, authentication, and platform restrictions.
Removing the flagged sources leaves $1{,}575/1{,}600$ MCQs, $302/316$ OEQs, and $513/546$ Grounding videos.
Table~\ref{tab:source_availability} gives the retained category counts.
These are availability-filtered subset sizes; the main results use the benchmark and effective sample counts specified in Table~\ref{tab:evaluation_coverage}.
% Source: NeurIPS rebuttal, RXIm0KA5jE / BJRVZeRIpr;
% verified against nips26_re/analysis/results/available_benchmark_by_task_type/.
\begin{table}[htbp]
\centering
\caption{Benchmark items retained after excluding the 50 sources that failed the July 2026 availability audit. Entries are retained/original counts. MCQ and OEQ counts refer to questions; Grounding counts refer to videos.}
\label{tab:source_availability}
\small
\setlength{\tabcolsep}{4pt}
\begin{tabular}{lrrrrrrr}
\toprule
Track & Audio & Social & Narrative & Perception & Scene & Total & Retained \\
\midrule
MCQ & 300/309 & 417/419 & 249/255 & 263/268 & 346/349 & 1575/1600 & 98.44\% \\
OEQ & 64/66 & 33/34 & 45/48 & 94/98 & 66/70 & 302/316 & 95.57\% \\
Grounding & -- & -- & -- & -- & -- & 513/546 & 93.96\% \\
\bottomrule
\end{tabular}
\end{table}

Source IDs/URLs, temporal boundaries, and retrieval and reconstruction instructions support reproducibility.

\subsection{Statistical Uncertainty}
\label{supp:stat_sig}
For the main MCQ comparisons, we report nonparametric 95\% bootstrap confidence intervals over test questions.
Each resample draws the same number of questions as the model's scored set with replacement and recomputes the overall accuracy.
We use 10,000 resamples and take the empirical 2.5th and 97.5th percentiles as the interval endpoints.
The intervals estimate finite-sample variability over the benchmark questions.
\begin{table}[t]\centering
\caption{95\% percentile bootstrap confidence intervals for MCQ accuracy, computed from 10,000 resamples of each model's scored questions.}
\label{tab:app_bootstrap_ci}
\setlength{\tabcolsep}{4.5pt}
\begin{tabular}{lccc}
\toprule
\textbf{Model} & \textbf{Scored questions} & \textbf{Accuracy} & \textbf{95\% CI} \\ \midrule
Phi-4-multimodal & 1598 & 24.22 & [22.15, 26.35] \\
Kimi-VL & 1600 & 22.56 & [20.56, 24.62] \\
InternVL3.5 & 1600 & 30.88 & [28.62, 33.12] \\
Ministral-3 & 1600 & 22.69 & [20.62, 24.69] \\
MiMo-VL & 1600 & 27.88 & [25.69, 30.12] \\
GLM-4.1V & 1600 & 23.25 & [21.12, 25.37] \\
GLM-4.6V-Flash & 1363 & 31.11 & [28.69, 33.60] \\
GLM-4.6V & 1600 & 32.50 & [30.25, 34.81] \\
Qwen3-VL (8B) & 1600 & 29.75 & [27.44, 32.06] \\
Qwen3-VL (32B) & 1600 & 33.19 & [30.94, 35.44] \\
Qwen2.5-VL & 1600 & 36.88 & [34.50, 39.19] \\
MiniCPM-o-2.6 & 1600 & 22.44 & [20.38, 24.50] \\
OmniVinci & 1597 & 29.49 & [27.24, 31.75] \\
Qwen2.5-Omni & 1596 & 24.12 & [22.06, 26.25] \\
Qwen3-Omni & 1599 & 38.34 & [36.02, 40.71] \\
MiniCPM-o-4.5 & 1575 & 31.30 & [29.02, 33.65] \\
MiMo-v2.5 & 1600 & 58.44 & [56.06, 60.88] \\
GPT-5 & 1600 & 63.50 & [61.12, 65.81] \\
Qwen3.5-Omni-Plus & 982 & 61.20 & [58.14, 64.15] \\
Gemini-3 Pro & 1574 & 67.66 & [65.37, 69.95] \\
\bottomrule
\end{tabular}
\end{table}

\begin{table}[t]
\centering
\caption{Effective sample counts for the main results. MCQ and OEQ counts are scored questions out of 1,600 and 316; $n_{\rm G}$ is parsed videos out of 546; $n_{\rm v}$ is parsed videos with annotated face boxes out of 540. -- denotes an unavailable count or unevaluated task.}
\label{tab:evaluation_coverage}
\setlength{\tabcolsep}{4.5pt}
\begin{tabular}{lrrrr}
\toprule
\textbf{Model} & $n_{\rm MCQ}$ & $n_{\rm OEQ}$ & $n_{\rm G}$ & $n_{\rm v}$ \\
\midrule
Phi-4-multimodal & 1598 & 315 & 79 & 79 \\
Kimi-VL & 1600 & 316 & -- & -- \\
InternVL3.5 & 1600 & 316 & -- & -- \\
Ministral-3 & 1600 & 316 & -- & -- \\
MiMo-VL & 1600 & 315 & -- & -- \\
GLM-4.1V & 1600 & 316 & -- & -- \\
GLM-4.6V-Flash & 1363 & 265 & -- & -- \\
GLM-4.6V & 1600 & 316 & -- & -- \\
Qwen3-VL (8B) & 1600 & 316 & -- & -- \\
Qwen3-VL (32B) & 1600 & 316 & -- & -- \\
Qwen2.5-VL & 1600 & 316 & -- & -- \\
MiniCPM-o-2.6 & 1600 & 316 & 5 & 5 \\
OmniVinci & 1597 & 304 & 320 & 315 \\
Qwen2.5-Omni & 1596 & 316 & 501 & 495 \\
Qwen3-Omni & 1599 & 316 & 433 & 429 \\
MiniCPM-o-4.5 & 1575 & 314 & 361 & 357 \\
MiMo-v2.5 & 1600 & 315 & 493 & 487 \\
GPT-5 & 1600 & 316 & -- & -- \\
Qwen3.5-Omni-Plus & 982 & -- & 245 & 243 \\
Gemini-3 Pro & 1574 & 305 & 546 & 540 \\
\bottomrule
\end{tabular}
\end{table}

\begin{table}[t]
\centering
\caption{Effective sample counts for Gemini-3 Pro and Qwen3-Omni modality and sampling ablations. Counts use the same populations as Table~\ref{tab:evaluation_coverage}.}
\label{tab:ablation_coverage}
\setlength{\tabcolsep}{4.5pt}
\begin{tabular}{lrrr}
\toprule
\textbf{Input} & $n_{\rm MCQ}$ & $n_{\rm G}$ & $n_{\rm v}$ \\
\midrule
\multicolumn{4}{l}{\textit{Gemini-3 Pro}} \\
A only & 1600 & 541 & 535 \\
A+V & 1574 & 546 & 540 \\
\midrule
\multicolumn{4}{l}{\textit{Qwen3-Omni}} \\
A+V (1 FPS) & 1599 & 433 & 429 \\
A+V (256 FPV) & 1599 & 440 & 436 \\
A+V (128 FPV) & 1599 & 464 & 459 \\
A+V (64 FPV) & 1600 & 461 & 457 \\
A+V (32 FPV) & 1596 & 348 & 345 \\
A+V (16 FPV) & 1600 & 466 & 460 \\
V only & 1600 & 82 & 78 \\
A only & 1600 & 484 & 479 \\
\bottomrule
\end{tabular}
\end{table}

\noindent\textbf{Qwen3.5-Omni-Plus coverage.}
For \texttt{qwen3.5-omni-plus-2026-03-15}, some requests to the official API were blocked by content moderation and returned \texttt{DataInspectionFailed}.
After excluding failed requests and outputs without a valid parsed answer, 982 of 1,600 MCQs and 245 of 546 grounding videos contribute to scoring; 243 grounding videos also have annotated face boxes for vIoU.
These scores are conditional on the retained subsets, whose content distribution may differ from the full benchmark, and should be interpreted together with coverage.
The OEQ score is retained from the original evaluation, and its effective sample count is unavailable.

\subsection{Supplementary Diagnostic Analyses}
\label{supp:diagnostic_details}

\begin{table*}[t]
  \centering
  \scriptsize
  \caption{Ablation Study on Visual Sampling Density and Modality. We evaluate Qwen3-Omni with varying frame counts and audio/video inputs. Effective sample counts are in Table~\ref{tab:ablation_coverage}.}
  \label{tab:main_ablation}
    \begin{tabular}{
      lcc|cccccc|ccc
    }
      \toprule
      \multirow{2}{*}{Visual} & \multirow{2}{*}{Audio} & \multirow{2}{*}{\makecell{Frames}} & \multicolumn{6}{c|}{\textbf{\textit{\trackb}}}  &
      \multicolumn{3}{c}{\textit{\textbf{\tracka}}}
      \\
      \cline{4-9} \cline{10-12} &
      &
      & \textbf{{Avg.}}
      & Narr.
      & Perc.
      & Scene
      & Soci.
      & Audi.
      & tIoU
      & vIoU
      & WER$\downarrow$ \\
       \hline
      \cmark & \cmark & 1 FPS & 38.34 & 30.98 & \textbf{47.57} & \textbf{37.54} & 37.47 & 38.51 & 49.25 & 3.18 & 75.79 \\
      \cmark & \cmark & 256 & 38.27 & 34.90 & 44.57 & 36.10 & 35.56 & 41.75 & 50.40 & 3.01 & 71.52 \\
      \cmark & \cmark & 128 & \textbf{39.40} & \textbf{36.47} & 44.57 & \textbf{37.54} & 36.99 & 42.72 & 48.20 & 3.49 & 74.93 \\
      \cmark & \cmark & 64 & 37.31 & 32.55 & 44.03 & 34.38 & 35.56 & 41.10 & 50.48 & 3.67 & 75.92 \\
      \cmark & \cmark & 32 & 38.78 & 32.68 & 46.62 & 34.96 & \textbf{38.04} & 42.39 & 52.91 & 3.63 & 76.42 \\
      \cmark & \cmark & 16 & 38.06 & 33.33 & 43.28 & 34.38 & 36.04 & \textbf{44.34} & 54.66 & \textbf{3.82} & 72.20 \\
      \hline
      \cmark & \xmark & 1 FPS & 30.44 & 26.27 & 25.00 & 33.81 & 30.79 & 34.30 & 26.51 & 1.54 & 97.99 \\
      \xmark & \cmark & - & 37.81 & 35.29 & 41.79 & 33.81 & 36.52 & 42.72 & \textbf{63.58} & 1.92 & \textbf{36.77} \\
      \bottomrule
    \end{tabular}
\end{table*}

Table~\ref{tab:app_joint_mcq_sanity} reports the same overlapping-clip median split as Table~\ref{tab:joint_grounding_reasoning}, but uses MCQ accuracy instead of OEQ scores.
The mixed MCQ trends reflect the broader evidence required by these questions: visual details, scene-presentation cues, and social cues are not fully captured by speech grounding metrics.

\begin{table}[h]
\centering
\caption{Grounding quality and MCQ accuracy on overlapping clips. Better and Worse denote the two halves ranked by each grounding metric, using the same procedure as Table~\ref{tab:joint_grounding_reasoning}.}
\label{tab:app_joint_mcq_sanity}
\setlength{\tabcolsep}{4.5pt}
\begin{tabular}{llcccc}
\toprule
\textbf{Model} & \textbf{Split metric} & \textbf{MCQs / videos} & \textbf{Better} & \textbf{Worse} & \textbf{Gap} \\
\midrule
\multirow{3}{*}{Gemini-3 Pro}
& tIoU $\uparrow$ & 166 / 144 & 73.81 & 78.05 & $-4.24$ \\
& vIoU $\uparrow$ & 164 / 142 & 75.32 & 75.86 & $-0.54$ \\
& WER $\downarrow$ & 166 / 144 & 76.47 & 75.31 & $+1.16$ \\
\midrule
\multirow{3}{*}{Qwen3-Omni}
& tIoU $\uparrow$ & 141 / 121 & 33.33 & 43.94 & $-10.61$ \\
& vIoU $\uparrow$ & 140 / 120 & 32.86 & 44.29 & $-11.43$ \\
& WER $\downarrow$ & 141 / 121 & 28.38 & 49.25 & $-20.88$ \\
\bottomrule
\end{tabular}
\end{table}

For both diagnostics, videos with valid predictions are sorted by grounding quality and divided at the middle rank; scores average valid questions within each group.
The OEQ tIoU/WER splits include 141 questions from 112 videos for Gemini-3 Pro and 104 questions from 88 videos for Qwen3-Omni.
After excluding videos without annotated face boxes, the vIoU splits use 138 questions from 110 videos and 102 questions from 87 videos, respectively.

Table~\ref{apptab:dataset_comparison} compares \bench~with representative video understanding benchmarks.
While many existing datasets focus on a single capability, such as multiple-choice reasoning, open-ended question answering, or isolated spatial or temporal grounding, \bench~combines \emph{multimodal reasoning (both MCQ and open-ended), spatio-temporal grounding, and speech understanding} within a unified audio--visual evaluation framework.
With manually verified and automatically scaled annotations, \bench~supports both structured QA and free-form perception, enabling a comprehensive assessment of MLLMs across fine-grained localization, temporal reasoning, and cross-modal audio--visual grounding.

% token + 人力
% 体现优势
% 每个类别三个代表
\begin{table*}[ht]
\centering
\scriptsize
\setlength{\tabcolsep}{4pt}
\caption{
 {Comparison between \bench~and existing video understanding benchmarks.} \bench~jointly supports reasoning, speech understanding, temporal grounding, and spatial grounding in the narrative-video domain. ``A'' means audio, ``V'' means video, ``I'' means image. ``Anno.'' is the annotation type (A: automatic, M: manual).}

\begin{tabular}{l|cccc|cc|ccc}
\toprule
& \multicolumn{4}{c|}{\textbf{}}  &  \multicolumn{2}{c|}{\textbf{Reasoning Q\&A}} & \multicolumn{3}{c}{\textbf{Free-Form Perception}} \\
 \textbf{Benchmark} & \textbf{\# Videos} & \textbf{\# Q\&A} &  \makecell{\textbf{Anno.}}  & \textbf{Modality} & 
\makecell{\textbf{MCQ}} & 
\makecell{\textbf{Open}} & 
\makecell{\textbf{Spatial-} \\ \textbf{Grounding}} & 
\makecell{\textbf{Temporal-} \\ \textbf{Grounding}} & \textbf{Speech}\\
\hline
\multicolumn{10}{l}{\violet{\textit{\textbf{Visual Benchmarks:}}}} \\
 \rowcolor{gray!10}
 \hline
 MV-Bench~\citep{liMVBenchComprehensiveMultimodal2024} & 3,641 & 4.0 K & A+M & V & \cmark & \xmark &  \xmark & \xmark & \xmark \\
 EgoSchema~\citep{mangalamEgoSchemaDiagnosticBenchmark2023}&  5,063 & 5.0 K & A+M & V & \cmark & \xmark &  \xmark & \xmark & \xmark \\
 \rowcolor{gray!10}
 MLVU~\citep{zhouMLVUBenchmarkingMultitask2025} & 1,730 & 3.1 K & A+M & V & \cmark & \cmark &  \xmark & \xmark & \xmark \\
 Video-MME~\citep{fuVideoMMEFirstEverComprehensive2025}  & 900 & 2.7 K & M & V & \cmark & \xmark &  \xmark & \xmark & \xmark \\
 \rowcolor{gray!10}
 LVBench~\citep{wangLVBenchExtremeLong2025}  & 103 & 1.6 K & M  & V & \cmark & \xmark &  \xmark & \xmark & \xmark \\
 \rowcolor{gray!10}
 LongVideoBench~\citep{wuLongVideoBenchBenchmarkLongcontext2024}  & 3,763 & 6.7 K & M & V & \cmark & \xmark &  \xmark & \xmark & \xmark \\
 VRBench~\citep{yuVRBenchBenchmarkMultiStep2025}  & 960 & 8.2 K & A+M & V & \cmark & \cmark &  \xmark & \xmark & \xmark \\
\hline
 \multicolumn{10}{l}{\violet{\textit{\textbf{Omni Benchmarks:}}}} \\
 \rowcolor{gray!10}
 \hline
 OmniBench~\citep{liOmniBenchFutureUniversal2025} & - & 1.1 K & A+M & I+A & \cmark & \xmark &  \xmark & \xmark & \xmark \\
 AV-Odyssey~\citep{gongAVOdysseyBenchCan2024}  & - & 4.6 K & A+M & I+A & \cmark & \xmark &  \xmark & \xmark & \xmark \\
 \rowcolor{gray!10}
 LongVALE~\citep{gengLongVALEVisionAudioLanguageEventBenchmark2025} & 8.4 K & - & A & V+A & \xmark & \xmark & \xmark & \cmark & \cmark \\
 \rowcolor{gray!10}
Daily-Omni~\citep{zhouDailyOmniAudioVisualReasoning2025}  & 684 & 1.2 K & A & V+A & \cmark & \xmark &  \xmark & \xmark & \xmark \\
 WorldSense~\citep{hongWorldSenseEvaluatingRealworld2025}  & 1,662 & 3.2 K & M & V+A & \cmark & \xmark &  \xmark & \xmark & \xmark \\
 \rowcolor{gray!10}
OmniVideoBench~\citep{liOmniVideoBenchAudioVisualUnderstanding2025}  & 628 & 1.0 K & M & V+A & \cmark & \xmark &  \xmark & \xmark & \xmark \\
\hline
\multicolumn{10}{l}{\violet{\textit{\textbf{Movie Benchmarks:}}}} \\
 MovieQA~\citep{tapaswiMovieQAUnderstandingStories2016}  & 140 & 6.5 K & M & V & \cmark & \xmark &  \xmark & \xmark & \xmark \\
\rowcolor{gray!10}
CinePile~\citep{rawalCinePileLongVideo2024}  &9396 & 304 K & A+M & V & \cmark & \xmark &  \xmark & \xmark & \xmark \\
 Movie101/v2~\citep{yueMovie101NewMovie2023,yueMovie101v2ImprovedMovie2025} & - & - & M & V & \xmark & \cmark & \xmark & \xmark & \xmark \\
 \rowcolor{gray!10}
 InfiniBench~\citep{ataallahInfiniBenchBenchmarkLarge2025}  & 1217 & 87.7 K & A+M & V & \cmark & \cmark &  \xmark & \xmark & \xmark \\
Video-Holmes~\citep{chengVideoHolmesCanMLLM2025a}  &270 & 1.8 K & A+M & V & \cmark & \xmark &  \xmark & \xmark & \xmark \\
\hline
\multicolumn{10}{l}{\violet{\textit{\textbf{Grounding Benchmarks:}}}} \\
\rowcolor{gray!10} TimeLens-Bench~\citep{zhangTimeLensRethinkingVideo2025}  & 4,279  & 6.5 K & M & V & \xmark & \xmark &  \xmark & \cmark & \xmark \\
SVAG-Bench~\citep{hannanSVAGBenchLargeScaleBenchmark2025}  & 688  & 19.6 K & A+M & V & \xmark & \xmark &  \cmark & \cmark & \xmark \\
\rowcolor{gray!10} OmniGround~\citep{gaoOmniGroundComprehensiveSpatioTemporal2025}  & 3,475  & 3.5 K & A+M & V & \xmark & \xmark &  \cmark & \cmark & \xmark \\
Know-Show~\citep{sugandhikaKnowShowBenchmarkingVideoLanguage2025}  & 750 & 2.5 K & M & V & \xmark & \xmark &  \cmark & \cmark & \xmark \\
 \hline
 \rowcolor{gray!10} \textbf{\bench~(ours)}  &975 & 1.9 K& A+M&  V+A & \cmark & \cmark & \cmark & \cmark & \cmark \\
 \bottomrule
 \end{tabular}%

 \label{apptab:dataset_comparison}
 % \vspace{-2ex}
 \end{table*}

\section{Configurations of Evaluated Models}
\label{supp:model_details}
\label{app:model_details}
Table~\ref{tab:multimodal_model_configuration} provides detailed configurations for the evaluated MLLMs, including model versions, modality support, deployment choices, and inference backends.
We primarily run local model inference on NVIDIA A100 80GB GPUs using vLLM.
MiMo-v2.5 has publicly released weights~\cite{xiaomi2026mimo25}; our main evaluations and supplementary controls use the API model identifier \texttt{xiaomi/mimo-v2.5}.
The main evaluations use temperature $0.7$, and supplementary MCQ controls use temperature $0$; both use top-$p=0.01$.
For GPT-5, the API used in our experiments did not expose these sampling parameters.
\begin{table*}[htbp]
\centering
\caption{Model identifiers and inference backends. Frame-sampling settings are in Table~\ref{tab:main_results}. HF denotes Hugging Face.}

  \resizebox{\linewidth}{!}{
  \label{tab:multimodal_model_configuration}
    \begin{tabular}{l c c c c c c }
      \toprule
      \multirow{2}{*}{\textbf{Models}} & 
      \multirow{2}{*}{\makecell{\textbf{Organization}}} &
      \multirow{2}{*}{\makecell{\textbf{Release}}} & 
      \multirow{2}{*}{\makecell{\textbf{Version}}} & 
      \multirow{2}{*}{\makecell{\textbf{Video}\\\textbf{Support}}} & 
      \multirow{2}{*}{\makecell{\textbf{Audio}\\\textbf{Support}}} & 
      \multirow{2}{*}{\makecell{\textbf{Inference}\\\textbf{Deployment}}} \\
       & & & & & & \\
      \midrule
      \tabtitlecolor
      \multicolumn{7}{c}{\textit{\textbf{Open-source / open-weight MLLMs}}} \\
      \midrule
      Phi-4-multimodal~\cite{microsoftPhi4MiniTechnicalReport2025} & Microsoft
      &2025-3& \texttt{Phi-4-multimodal-instruct} & \xmark & \cmark & vLLM \\
      Kimi-VL~\cite{teamKimiVLTechnicalReport2025} & Moonshot  
      &2025-4& \texttt{Kimi-VL-A3B-Thinking} & \xmark & \xmark & vLLM \\
      InternVL3.5~\cite{wangInternVL35AdvancingOpenSource2025}  & OpenGVLab 
      & 2025-8& \texttt{InternVL3\_5-38B-Instruct} & \xmark & \xmark & vLLM \\
      Ministral-3~\cite{liuMinistral32026} & Mistral 
      &2025-12& \texttt{Ministral-3-14B-Reasoning-2512} & \xmark & \xmark & vLLM \\
      MiMo-VL~\cite{liXiaomiMiMoVLMilocoTechnical2025} & Xiaomi  
      &2025-6& \texttt{MiMo-VL-7B-RL-2508} & \cmark & \xmark & vLLM \\
      GLM-4.1V~\cite{teamGLM45VGLM41VThinkingVersatile2025} & Zhipu AI 
      &2025-7& \texttt{GLM-4.1V-9B-Base} & \cmark & \xmark & vLLM \\
      GLM-4.6V-Flash~\cite{teamGLM45VGLM41VThinkingVersatile2025} &Zhipu AI
      &2025-12& \texttt{GLM-4.6V-Flash} & \cmark & \xmark & SGlang \\
      GLM-4.6V~\cite{teamGLM45VGLM41VThinkingVersatile2025} &Zhipu AI 
      &2025-12& \texttt{GLM-4.6V} & \cmark & \xmark & SGLang \\
      Qwen3-VL (8B)~\cite{baiQwen3VLTechnicalReport2025} &Alibaba 
      &2025-10 & \texttt{Qwen3-VL-8B-Instruct} & \cmark & \xmark & vLLM \\
      Qwen3-VL (32B)~\cite{baiQwen3VLTechnicalReport2025} &Alibaba
      &2025-10&\texttt{ Qwen3-VL-32B-Instruct} & \cmark & \xmark & vLLM \\
      Qwen2.5-VL (72B)~\cite{baiQwen25VLTechnicalReport2025} &Alibaba
      &2025-2& \texttt{Qwen2.5-VL-72B-Instruct} & \cmark & \xmark & vLLM \\
      MiniCPM-o-2.6~\cite{yaoMiniCPMVGPT4VLevel2024} & Openbmb 
      &2025-1& \texttt{MiniCPM-o-2\_6} & \cmark & \cmark & HF \\
      Omnivinci~\cite{yeOmniVinciEnhancingArchitecture2025} &Nvidia 
      &2025-10& \texttt{nvidia/omnivinci} & \cmark & \cmark & HF \\
      Qwen2.5-Omni~\cite{xuQwen25OmniTechnicalReport2025} &Alibaba  
      &2025-3& \texttt{Qwen2.5-Omni-7B} & \cmark & \cmark & vLLM \\
      Qwen3-Omni~\cite{xuQwen3OmniTechnicalReport2025} & Alibaba 
      &2025-9& \texttt{Qwen3-Omni-30B-A3B-Instruct} & \cmark & \cmark & vLLM \\
      MiniCPM-o-4.5~\cite{cui2026minicpmo45realtimefullduplex} & Openbmb 
      &2026-4& \texttt{MiniCPM-o-4\_5} & \cmark & \cmark & vLLM \\
      MiMo-v2.5~\cite{xiaomi2026mimo25} & Xiaomi
      & -- & \texttt{xiaomi/mimo-v2.5} & \cmark & \cmark & API \\
      \midrule
      \tabtitlecolor \multicolumn{7}{c}{\textit{\textbf{Closed-source MLLMs}}} \\
      \midrule
      GPT-5~\cite{OpenAI2025GPT5SystemCard} & OpenAI 
      &2025-8& \texttt{GPT-5-2025-08-07} & \xmark & \xmark & API \\
      Gemini-3 Pro~\cite{gemini-3-pro} & Google 
      &2025-11& \texttt{Gemini-3-pro-preview} & \cmark & \cmark & API \\
      Qwen3.5-Omni-Plus~\cite{qwenteam2026qwen35omnitechnicalreport} & Alibaba
      &2026-04& \texttt{qwen3.5-omni-plus-2026-03-15} & \cmark & \cmark & API \\
      \bottomrule
    \end{tabular}}
\end{table*}

\subsection{Prompt for MLLMs}

% Auto-generated by Python script
% Title: MCQ_COT_PROMPT
\begin{tcolorbox}[
    breakable,
    colback=gray!5, 
    colframe=gray!60!black,
    title=\textbf{MCQ-CoT-PROMPT}, 
    fontupper=\small\ttfamily,
    boxrule=0.5mm,
    arc=2mm,
    bottom=2mm, top=2mm
]

You are a helpful video understanding assistant that answers multi-choice questions through step-by-step reasoning based on the video.

\# Instructions:

1. Break down the reasoning process into clear, specific events.

2. Conclude with the best option letter (A/B/C/D/E/F) at last.

\# Output Format:

\textless{}Step 1\textgreater{} Description of event/observation 

\textless{}Step 2\textgreater{} Description of event/observation 

...  

\textless{}Answer\textgreater{} Option letter

\# Multiple Choice Question

\$multiple\_choice\_question

\end{tcolorbox}

% Auto-generated by Python script
% Title: OPENEND_PROMPT_R1
\begin{tcolorbox}[
    breakable,
    colback=gray!5, 
    colframe=gray!60!black,
    title=\textbf{OEQ-CoT-PROMPT}, 
    fontupper=\small\ttfamily,
    boxrule=0.5mm,
    arc=2mm,
    bottom=2mm, top=2mm
]

You are a helpful video understanding assistant that answers Open-end questions through step-by-step reasoning based on the video.

\# Instructions:

1. Break down the reasoning process into clear, specific events.

2. Conclude with the best answer at last.

\# Output Format:

\textless{}Step 1\textgreater{} Description of event/observation 

\textless{}Step 2\textgreater{} Description of event/observation 

...  

\textless{}Answer\textgreater{} Your Answer

\# Open End Question

\$open\_end\_question

\end{tcolorbox}

\section{Structured-Output Robustness for Grounding Track}
\label{app:grounding_exclusions}

The \tracka~protocol requires machine-checkable structured outputs containing speech transcripts, temporal boundaries, and speaker face boxes.
This requirement is part of the benchmark rather than a mere engineering detail: long-form grounding systems must not only perceive evidence but also report it in a reproducible format.
Table~\ref{tab:open_closed_llm_method} includes early omni-modal models such as \textbf{Omnivinci}~\citep{yeOmniVinciEnhancingArchitecture2025} and \textbf{MiniCPM-o-2.6}~\citep{yaoMiniCPMVGPT4VLevel2024}.
The reported grounding metrics average over successfully parsed videos; missing or unparseable predictions are excluded from these means.
Table~\ref{tab:evaluation_coverage} reports the number of parsed videos for each model, with vIoU additionally restricted to videos with annotated face boxes.
The main sources of failure are:

\begin{itemize}
    \item \textbf{Engineering framework limitations:} Limited support for constrained decoding can lead to malformed outputs or missing required fields.
    \item \textbf{Instruction-following limitations:} Complex grounding schemas require consistent timestamp, transcript, and box formatting. Several models revert to free-text descriptions or produce partially valid JSON, which prevents reliable automated alignment.
\end{itemize}

\section{Qualitative Analysis and Case Studies}
\label{supp:qualitative_examples}

To provide a concrete understanding of the reasoning depth required by CineMME, we showcase detailed qualitative examples across the five core dimensions of the \textit{\trackb}. Each case study illustrates a specific query type, the corresponding candidate options, and the Chain-of-Thought (CoT) reasoning traces generated by state-of-the-art MLLMs (e.g., Gemini-3 Pro~\citep{gemini-3-pro}, Qwen3-Omni~\citep{xuQwen3OmniTechnicalReport2025}).

\begin{figure}[t]
    \centering
    
    \includegraphics[width=0.85\textwidth]{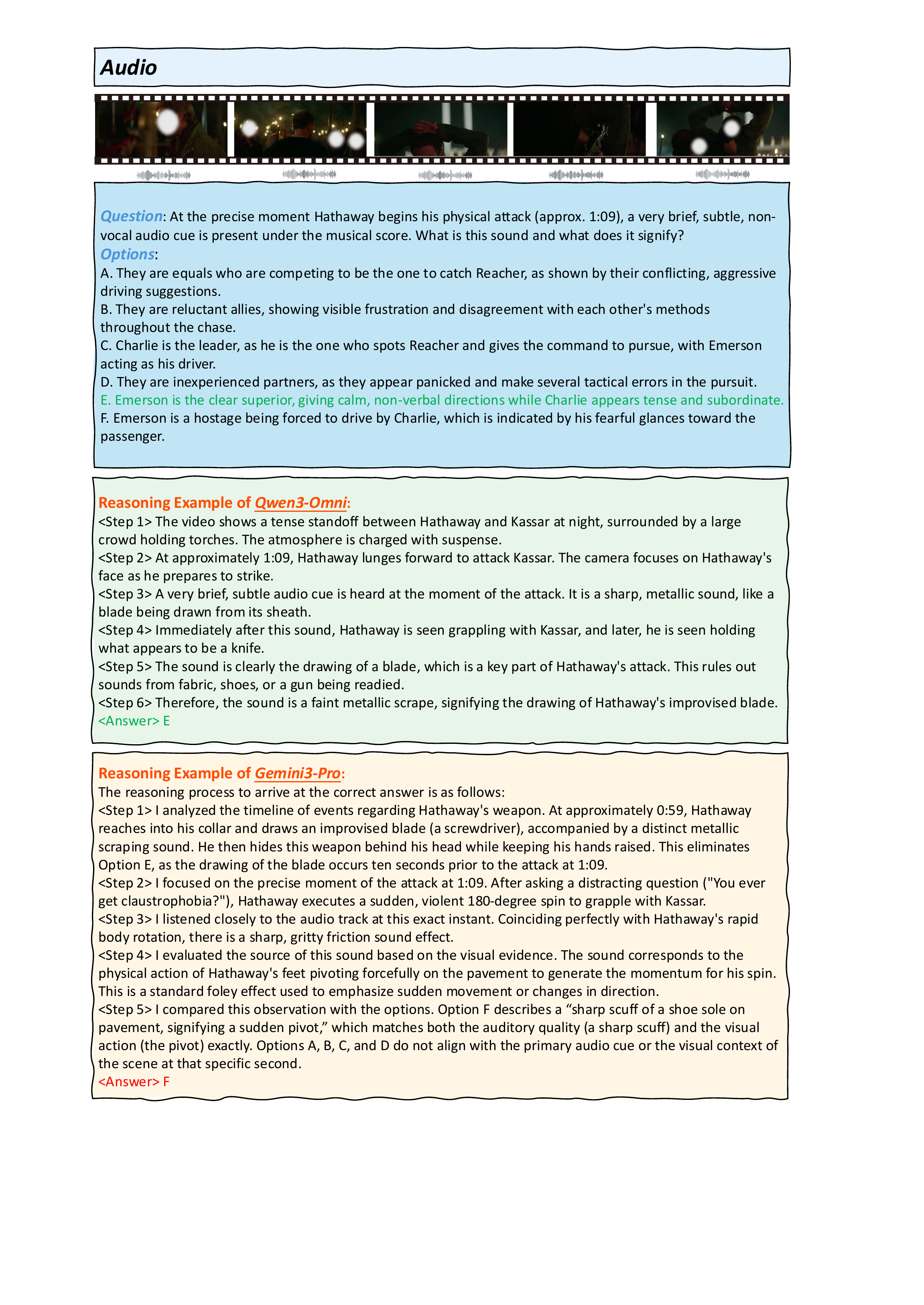}
    
    \caption{An example of an MCQ in the \textbf{Audio} category of a \trackb~ shows the Question, Options, and highlights the correct answer in green. Below, an example of correct Reasoning and an example of incorrect Reasoning is illustrated.}
    \label{fig:qa_case_Audio}
\end{figure}

\begin{figure}[t]
    \centering
    
    \includegraphics[width=\textwidth]{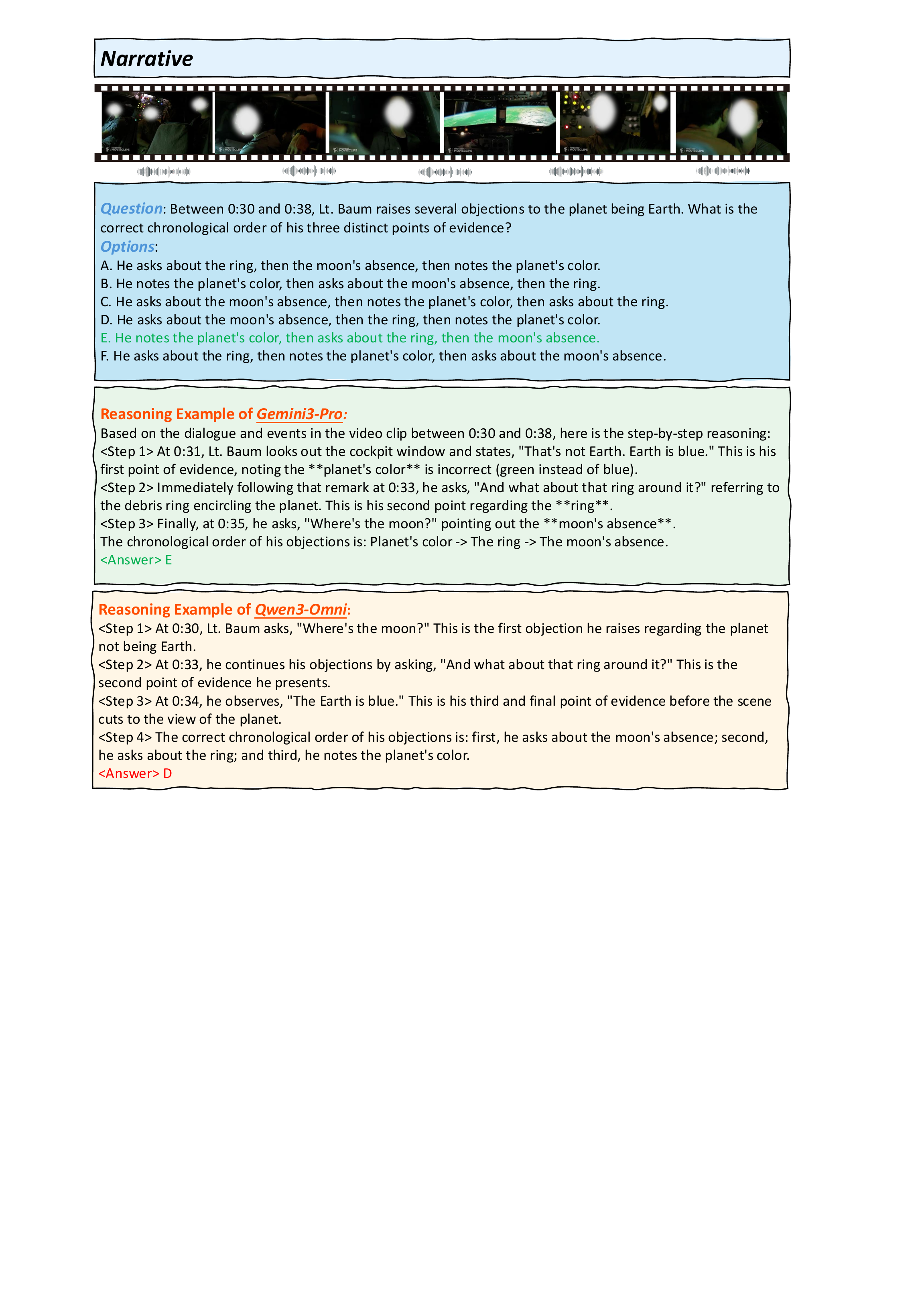}
    \caption{An example of an MCQ in the \textbf{Narrative} category of a \trackb~ shows the Question, Options, and highlights the correct answer in green. Below, an example of correct Reasoning and an example of incorrect Reasoning is illustrated.}
    \label{fig:qa_case_Narrative}
\end{figure}

\begin{figure}[t]
    \centering
    
    \includegraphics[width=\textwidth]{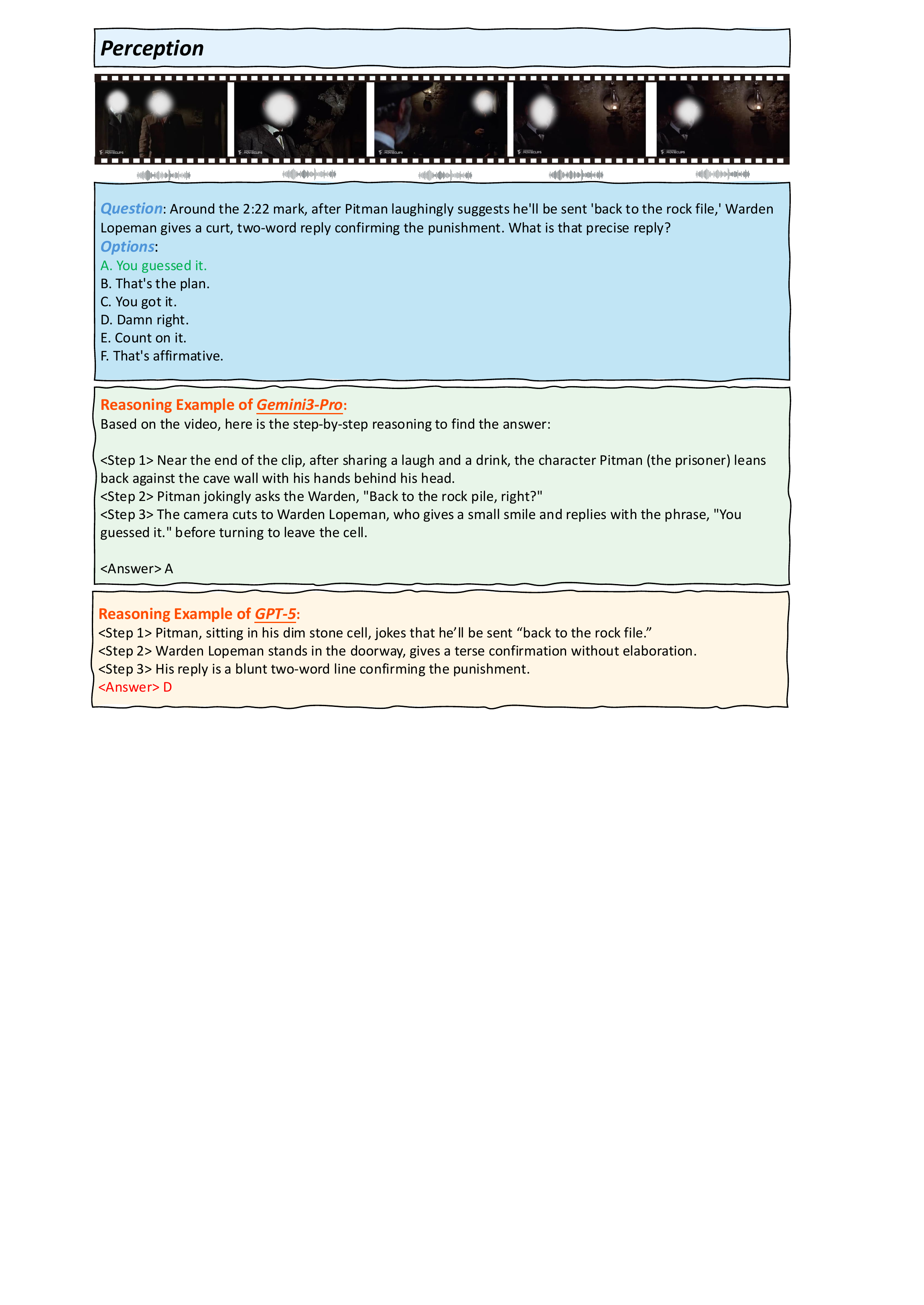}
    
    \caption{An example of an MCQ in the \textbf{Perception} category of a \trackb~ shows the Question, Options, and highlights the correct answer in green. Below, an example of correct Reasoning and an example of incorrect Reasoning is illustrated.}
    \label{fig:qa_case_Perception}
\end{figure}

\begin{figure}[t]
    \centering
    
    \includegraphics[width=\textwidth]{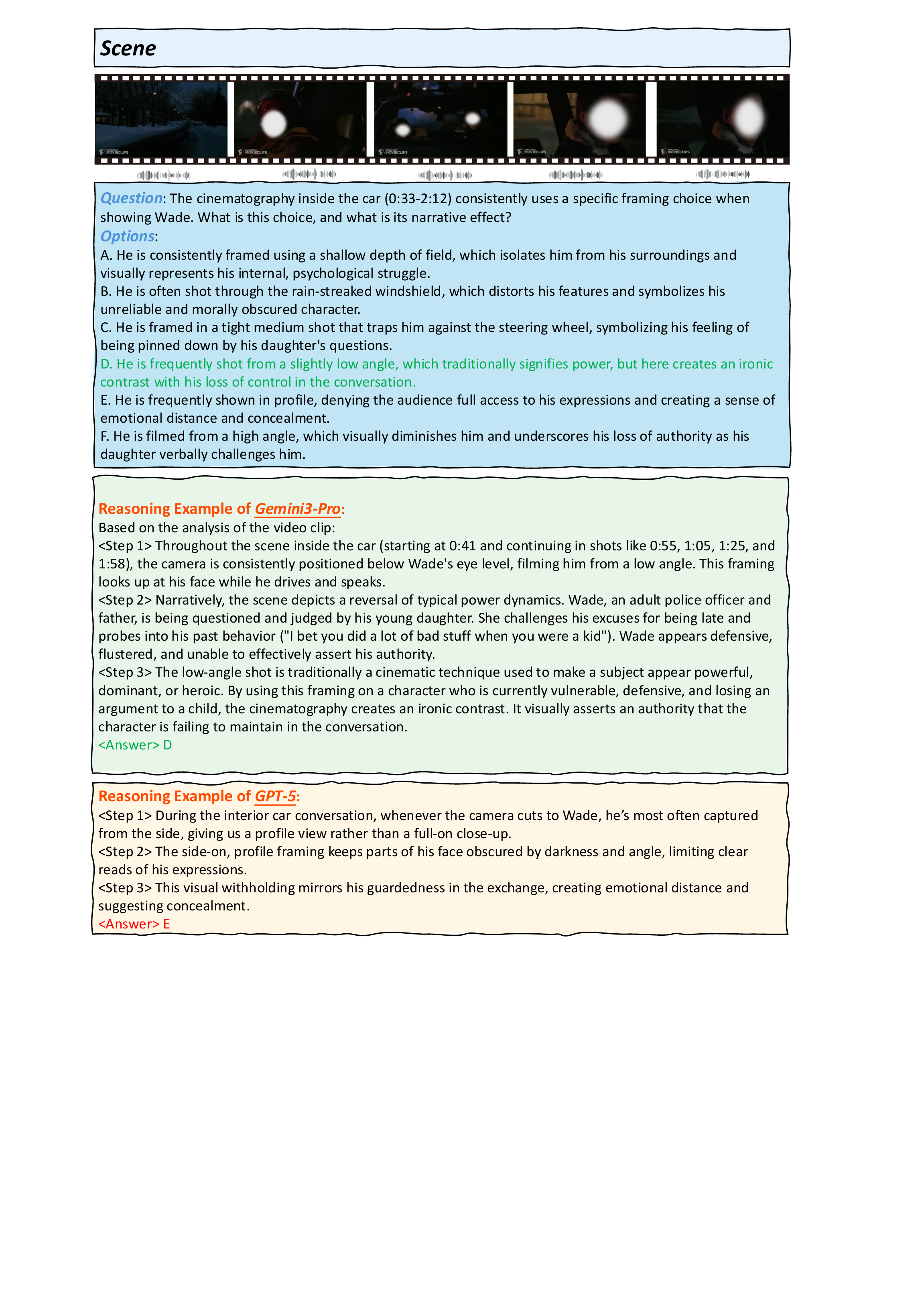}
    
    \caption{An example of an MCQ in the \textbf{Scene Presentation} category of a \trackb~ shows the Question, Options, and highlights the correct answer in green. Below, an example of correct Reasoning and an example of incorrect Reasoning is illustrated.}
    \label{fig:qa_case_scene_presentation}
\end{figure}

\begin{figure}[t]
    \centering
    
    \includegraphics[width=\textwidth]{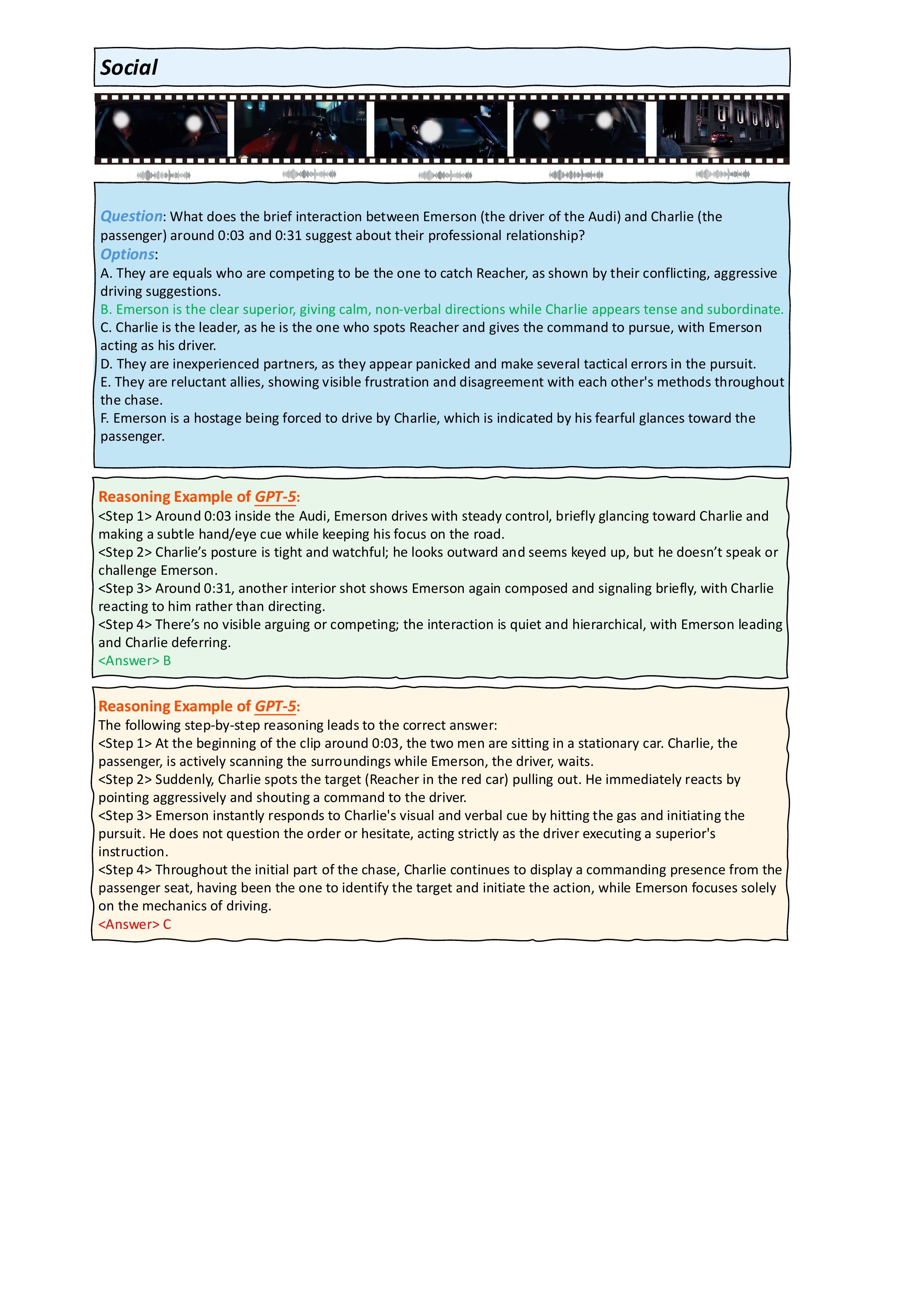}
    
    \caption{An example of an MCQ in the \textbf{Social} category of a \trackb~ shows the Question, Options, and highlights the correct answer in green. Below, an example of correct Reasoning and an example of incorrect Reasoning is illustrated.}
    \label{fig:qa_case_Social}
\end{figure}

These examples highlight distinct capabilities and common failure modes:

\begin{itemize}
    \item \textbf{Narrative:} This domain tests the model's ability to track and order events within a dialogue-heavy stream. As shown in Figure \ref{fig:qa_case_Narrative}, the model must correctly sequence a character's logical objections (e.g., distinguishing the order of remarks on ``planet color,'' ``rings,'' and ``moon''). We observe that while some models capture the content, they often hallucinate the chronological timeline of the arguments.
    
    \item \textbf{Perception:} Evaluates the detection of fleeting visual cues and specific character responses that require precise frame-level grounding, as shown in Figure \ref{fig:qa_case_Perception}.
    
    \item \textbf{Audio:} Demonstrates the necessity of distinguishing subtle acoustic events. as shown in Figure \ref{fig:qa_case_Audio}, models must differentiate between a ``metallic scrape'' (signifying a weapon) and a ``shoe scuff'' (signifying movement) to correctly interpret the narrative context.
    
    \item \textbf{Scene Presentation \& Social:} Assesses the interpretation of visual composition cues (e.g., camera angles implying power dynamics as shown in Figure \ref{fig:qa_case_scene_presentation}) and the inference of implicit professional relationships between characters as shown in Figure \ref{fig:qa_case_Social}.
\end{itemize}

These cases confirm that high performance on \bench requires holistic reasoning that aligns visual dynamics, acoustic cues, and chronological logic.

%%%%%%%%%%%%%%%%%%%%%%%%%%%%%%%%%%%%%%%%%%%%%%%%%%%%%%%%%%%%

\clearpage
\newpage
\newpage
\newpage
\newpage
\section*{NeurIPS Paper Checklist}

%%% BEGIN INSTRUCTIONS %%%
The checklist is designed to encourage best practices for responsible machine learning research, addressing issues of reproducibility, transparency, research ethics, and societal impact. Do not remove the checklist: {\bf The papers not including the checklist will be desk rejected.} The checklist should follow the references and follow the (optional) supplemental material.  The checklist does NOT count towards the page
limit.

Please read the checklist guidelines carefully for information on how to answer these questions. For each question in the checklist:
\begin{itemize}
    \item You should answer \answerYes{}, \answerNo{}, or \answerNA{}.
    \item \answerNA{} means either that the question is Not Applicable for that particular paper or the relevant information is Not Available.
    \item Please provide a short (1--2 sentence) justification right after your answer (even for \answerNA).
   % \item {\bf The papers not including the checklist will be desk rejected.}
\end{itemize}

{\bf The checklist answers are an integral part of your paper submission.} They are visible to the reviewers, area chairs, senior area chairs, and ethics reviewers. You will also be asked to include it (after eventual revisions) with the final version of your paper, and its final version will be published with the paper.

The reviewers of your paper will be asked to use the checklist as one of the factors in their evaluation. While \answerYes{} is generally preferable to \answerNo{}, it is perfectly acceptable to answer \answerNo{} provided a proper justification is given (e.g., error bars are not reported because it would be too computationally expensive'' or ``we were unable to find the license for the dataset we used''). In general, answering \answerNo{} or \answerNA{} is not grounds for rejection. While the questions are phrased in a binary way, we acknowledge that the true answer is often more nuanced, so please just use your best judgment and write a justification to elaborate. All supporting evidence can appear either in the main paper or the supplemental material, provided in appendix. If you answer \answerYes{} to a question, in the justification please point to the section(s) where related material for the question can be found.

IMPORTANT, please:
\begin{itemize}
    \item {\bf Delete this instruction block, but keep the section heading ``NeurIPS Paper Checklist"},
    \item  {\bf Keep the checklist subsection headings, questions/answers and guidelines below.}
    \item {\bf Do not modify the questions and only use the provided macros for your answers}.
\end{itemize}

%%% END INSTRUCTIONS %%%

\begin{enumerate}

\item {\bf Claims}
    \item[] Question: Do the main claims made in the abstract and introduction accurately reflect the paper's contributions and scope?
    \item[] Answer: \answerYes{} % Replace by \answerYes{}, \answerNo{}, or \answerNA{}.
    \item[] Justification: The abstract and introduction state claims that are directly supported by the dataset statistics, annotation verification results, benchmark comparisons, and experiments reported in the main paper and appendix. We frame CineMME as a diagnostic benchmark for narrative omni-modal understanding rather than as a deployed model or complete solution, and the main claims are scoped to the released dataset, evaluation protocols, and evaluated MLLMs. Limitations and responsible-use considerations are discussed in Section~\ref{sec:discussion}.

\item {\bf Limitations}
    \item[] Question: Does the paper discuss the limitations of the work performed by the authors?
    \item[] Answer: \answerYes{} % Replace by \answerYes{}, \answerNo{}, or \answerNA{}.
    \item[] Justification: Section~\ref{sec:discussion} and Appendix~\ref{supp:societal_impact} discuss dataset scope, copyright and release constraints, cultural/domain bias, grounding/reasoning coverage, LLM-as-a-judge evaluation noise, hosted API reproducibility limits, and responsible-use considerations for face-box annotations.

\item {\bf Theory assumptions and proofs}
    \item[] Question: For each theoretical result, does the paper provide the full set of assumptions and a complete (and correct) proof?
    \item[] Answer: \answerNA{} % Replace by \answerYes{}, \answerNo{}, or \answerNA{}.
    \item[] Justification: The paper does not present theoretical results, theorems, or formal proofs. It is an empirical benchmark paper focused on dataset construction, evaluation protocols, and experimental analysis.

    \item {\bf Experimental result reproducibility}
    \item[] Question: Does the paper fully disclose all the information needed to reproduce the main experimental results of the paper to the extent that it affects the main claims and/or conclusions of the paper (regardless of whether the code and data are provided or not)?
    \item[] Answer: \answerYes{} % Replace by \answerYes{}, \answerNo{}, or \answerNA{}.
    \item[] Justification: We provide benchmark metadata and annotations, source URLs and timestamps, evaluation protocols, metrics, prompts, model configurations, parsed outputs, and evaluation scripts for reproducing or verifying the main MCQ, OEQ, and grounding results. Open-source model checkpoints and inference settings are documented; for hosted API evaluations, we provide prompts, settings, outputs, and scoring code, while noting that exact regeneration may depend on API availability and model-version changes.

\item {\bf Open access to data and code}
    \item[] Question: Does the paper provide open access to the data and code, with sufficient instructions to faithfully reproduce the main experimental results, as described in supplemental material?
    \item[] Answer: \answerYes{} % Replace by \answerYes{}, \answerNo{}, or \answerNA{}.
    \item[] Justification: We provide a GitHub repository with the benchmark metadata, annotations, prompts, preprocessing utilities, evaluation code, and instructions for reproducing the main MCQ, OEQ, and grounding results. The repository includes environment setup, dataset format documentation, example commands, and scripts for running evaluation from model predictions. Source URLs, video identifiers, timestamps, and derived annotations are accompanied by data-preparation instructions, with source videos subject to the original platform and copyright terms.

\item {\bf Experimental setting/details}
    \item[] Question: Does the paper specify all the training and test details (e.g., data splits, hyperparameters, how they were chosen, type of optimizer) necessary to understand the results?
    \item[] Answer: \answerYes{} % Replace by \answerYes{}, \answerNo{}, or \answerNA{}.
    \item[] Justification: The main paper describes the CineMME-Reasoning and CineMME-Grounding evaluation protocols, data splits, input modalities, and metrics. Appendix~\ref{supp:model_details} and Table~\ref{tab:multimodal_model_configuration} provide model-specific deployment settings, decoding parameters, frame/audio sampling settings, and API or local inference configurations. Appendix~\ref{sec:appendix_eval_prompt} provides the prompts used for MCQ evaluation, open-ended judging, adversarial refinement, and grounding evaluation. No model training or optimizer selection is performed in this work.

\item {\bf Experiment statistical significance}
    \item[] Question: Does the paper report error bars suitably and correctly defined or other appropriate information about the statistical significance of the experiments?
    \item[] Answer: \answerYes{} % Replace by \answerYes{}, \answerNo{}, or \answerNA{}.
    \item[] Justification:  Appendix~\ref{supp:stat_sig} defines the nonparametric bootstrap protocol for the main MCQ reasoning comparisons, and Table~\ref{tab:app_bootstrap_ci} reports the $95\%$ confidence intervals and scored sample counts. Section~\ref{sec:main_results} discusses the main comparisons and refers to these intervals.

\item {\bf Experiments compute resources}
    \item[] Question: For each experiment, does the paper provide sufficient information on the computer resources (type of compute workers, memory, time of execution) needed to reproduce the experiments?
    \item[] Answer: \answerYes{}
    \item[] Justification: Local model inference primarily uses NVIDIA A100 80GB GPUs with vLLM; MiMo-v2.5 and proprietary models are evaluated via APIs. Appendix~\ref{supp:model_details} and Table~\ref{tab:multimodal_model_configuration} report model versions, inference backends, and decoding settings.
    
\item {\bf Code of ethics}
    \item[] Question: Does the research conducted in the paper conform, in every respect, with the NeurIPS Code of Ethics \url{https://neurips.cc/public/EthicsGuidelines}?
    \item[] Answer: \answerYes{} % Replace by \answerYes{}, \answerNo{}, or \answerNA{}.
    \item[] Justification: We have reviewed the NeurIPS Code of Ethics. CineMME is an evaluation benchmark built from publicly available narrative video clips; we release video identifiers, timestamps, annotations, prompts, evaluation code, and model outputs. We discuss copyright, face-box annotation risks, cultural/domain bias, misuse prevention, release safeguards, and benchmark scope in Section~\ref{sec:discussion} and the Societal Impact section.

\item {\bf Broader impacts}
    \item[] Question: Does the paper discuss both potential positive societal impacts and negative societal impacts of the work performed?
    \item[] Answer: \answerYes{} % Replace by \answerYes{}, \answerNo{}, or \answerNA{}.
    \item[] Justification: The paper discusses positive and negative societal impacts in Appendix~\ref{supp:societal_impact} and Section~\ref{sec:discussion}. Positively, CineMME provides a reproducible diagnostic benchmark for narrative omni-modal understanding. The paper also discusses copyright-sensitive video use, privacy and misuse risks from face-box annotations, cultural/domain bias, and the risk of overinterpreting benchmark scores. Mitigations include documenting source provenance, documenting dataset scope and release policy, discouraging identity-recognition or surveillance uses, and providing transparent evaluation protocols.
    
\item {\bf Safeguards}
    \item[] Question: Does the paper describe safeguards that have been put in place for responsible release of data or models that have a high risk for misuse (e.g., pre-trained language models, image generators, or scraped datasets)?
    \item[] Answer: \answerYes{}
    \item[] Justification: The paper describes safeguards for responsible benchmark release in Appendix~\ref{supp:societal_impact} and Section~\ref{sec:discussion}. The benchmark resources include video identifiers, source URLs, timestamps, annotations, prompts, evaluation code, and model outputs under a documented research-use policy. To reduce misuse risk, the released annotations avoid identity labels and are intended for narrative video understanding and grounding evaluation rather than face recognition, surveillance, or person identification. The paper also documents copyright constraints, dataset scope, cultural/domain bias, and recommended use restrictions.

\item {\bf Licenses for existing assets}
    \item[] Question: Are the creators or original owners of assets (e.g., code, data, models), used in the paper, properly credited and are the license and terms of use explicitly mentioned and properly respected?
    \item[] Answer: \answerYes{}
\item[] Justification: The paper credits the existing assets used in the benchmark and experiments, including source video platforms, evaluated models, baseline benchmarks, and open-source tools. Original videos remain under the copyright and terms of their respective owners and platforms. We release video identifiers, source URLs, timestamps, annotations, prompts, evaluation code, and model outputs for research use. The released benchmark package includes license and terms-of-use information for our annotations and code, and documents that users should obtain any source videos in compliance with the original platform and copyright terms.

\item {\bf New assets}
    \item[] Question: Are new assets introduced in the paper well documented and is the documentation provided alongside the assets?
    \item[] Answer: \answerYes{}
    \item[] Justification: The paper introduces new benchmark assets, including CineMME annotations, prompts, evaluation code, metadata, and model outputs. These assets are documented in the paper, Appendix~\ref{supp:societal_impact}, Section~\ref{sec:discussion}, and the release package, which includes dataset structure, field definitions, evaluation instructions, license and terms-of-use information, and known limitations. We provide source URLs, video identifiers, timestamps, and derived annotations for research use. Appendix~\ref{supp:data_access} reports the source-availability audit and retained subsets.

\item {\bf Crowdsourcing and research with human subjects}
    \item[] Question: For crowdsourcing experiments and research with human subjects, does the paper include the full text of instructions given to participants and screenshots, if applicable, as well as details about compensation (if any)? 
    \item[] Answer: \answerYes{} % Replace by \answerYes{}, \answerNo{}, or \answerNA{}.
    \item[] Justification: The paper involves human annotation and verification for benchmark construction, as well as a small human scoring study for validating the LLM-as-a-judge protocol and a non-expert MCQ reference study (Appendix~\ref{supp:human_performance}). The main paper reports the annotation scale and verification statistics, and the appendix provides annotation effort, agreement between annotation and review stages, final verification outcomes and segment discard rates, human-judge consistency, task instructions, and compensation policy. We omit platform screenshots because they may contain platform-specific or identifying information, but provide representative task descriptions and anonymized instruction text sufficient to reproduce the annotation protocol.

\item {\bf Institutional review board (IRB) approvals or equivalent for research with human subjects}
    \item[] Question: Does the paper describe potential risks incurred by study participants, whether such risks were disclosed to the subjects, and whether Institutional Review Board (IRB) approvals (or an equivalent approval/review based on the requirements of your country or institution) were obtained?
    \item[] Answer: \answerYes{}
    \item[] Justification: The study involved human annotation and verification for benchmark construction and a small human scoring study. The annotation protocol was reviewed under the authors' institutional or equivalent ethics procedure and was approved or determined to be exempt from full IRB review. Participant risks were minimal and limited to viewing publicly available narrative video clips and providing annotation or scoring judgments. Annotators were informed of the task requirements, no personal or sensitive information about annotators was collected for research analysis, and compensation followed the annotation platform contract and local labor requirements.

\item {\bf Declaration of LLM usage}
    \item[] Question: Does the paper describe the usage of LLMs if it is an important, original, or non-standard component of the core methods in this research? Note that if the LLM is used only for writing, editing, or formatting purposes and does \emph{not} impact the core methodology, scientific rigor, or originality of the research, declaration is not required.
    %this research? 
    \item[] Answer: \answerYes{}
    \item[] Justification: The paper describes the use of LLMs and MLLMs as part of the benchmark construction and evaluation pipeline. Specifically, LLMs/MLLMs are used for candidate question generation, adversarial refinement of distractors, open-ended answer judging, and as evaluated models in the benchmark experiments. The paper and appendix provide the relevant prompts, evaluation protocol, human verification procedure, and human-judge consistency analysis.

\end{enumerate}

\end{document}